%% file: pfpe.tex
\documentclass{article}

\usepackage[accepted]{icml2023}

\usepackage[utf8]{inputenc} 
\usepackage[T1]{fontenc}    
\usepackage[backref]{hyperref}       
\usepackage{url}            
\usepackage{booktabs}       
\usepackage{amsfonts}       
\usepackage{nicefrac}       
\usepackage{microtype}      

\usepackage{amsmath} 
\usepackage{amssymb}  

\usepackage{times}
\usepackage{helvet}
\usepackage{courier}
\usepackage{url}
\usepackage{blkarray}

\usepackage{amsthm}
\usepackage[font=small]{caption}
\usepackage{natbib}
\usepackage[capitalize,noabbrev]{cleveref}
\usepackage{autonum}
\usepackage{mathtools}
\usepackage{siunitx}
\usepackage[colorinlistoftodos]{todonotes}
\usepackage{enumitem}
\usepackage{siunitx}
\usepackage{dsfont}
\usepackage[english]{babel}
\usepackage[parfill]{parskip}
\usepackage{bm}
\usepackage{mathrsfs}
\usepackage{multicol}
\usepackage{relsize}
\usepackage{nicefrac}

\usepackage{pdflscape}
\usepackage{afterpage}
\usepackage{graphicx}
\usepackage{makecell}
\usepackage{array}
\usepackage{subcaption}
\usepackage{dblfloatfix}
\usepackage{float}
\usepackage{tablefootnote}
\usepackage[flushleft]{threeparttable}
\usepackage{multirow}
\usepackage{wrapfig}
\graphicspath{{Figures/}}

\usepackage{algorithm}
\usepackage{algorithmic}
\usepackage{changepage}
\usepackage{bbm}

\newtheorem{assumption}{Assumption}

\newtheorem{theorem}{Theorem}
\newtheorem{lemma}{Lemma}
\newtheorem{corollary}{Corollary}[theorem]

\newtheorem{definition}{Definition}
\newtheorem{proposition}{Proposition}

\DeclareMathOperator*{\argsup}{arg\,sup}
\DeclareMathOperator*{\arginf}{arg\,inf}

\DeclareMathOperator*{\argmin}{arg\,min}
\DeclarePairedDelimiterX{\infdivx}[2]{(}{)}{%
	#1\;\delimsize\|\;#2%
}

\usepackage{tikz}
\usetikzlibrary{automata} 
\usetikzlibrary{positioning} 
\usetikzlibrary{arrows} 

\usepackage{comment}

\icmltitlerunning{Why Target Networks Stabilise Temporal Difference Methods}

\begin{document}

\twocolumn[
\icmltitle{Why Target Networks Stabilise Temporal Difference Methods
}



\icmlsetsymbol{equal}{*}

\begin{icmlauthorlist}
\icmlauthor{Mattie Fellows}{equal,yyy}
\icmlauthor{Matthew J.A. Smith}{equal,yyy}
\icmlauthor{Shimon Whiteson}{yyy}
\end{icmlauthorlist}

\icmlaffiliation{yyy}{Department of Computer Science, University of Oxford,Oxford, United Kingdom}

\icmlcorrespondingauthor{Mattie Fellows}{matthew.fellows@cs.ox.ac.uk}

\icmlkeywords{Machine Learning, ICML}

\vskip 0.3in
]



\printAffiliationsAndNotice{\icmlEqualContribution} 
	
	\input{Sections/00_abstract.tex}
	\input{Sections/01_introduction.tex}

	\input{Sections/02_preliminaries.tex}
	\input{Sections/03_pfpe.tex}
	\input{Sections/04_asymptotic_analysis.tex}
	\input{Sections/05_finite_analysis.tex}

	\input{Sections/06_related_work.tex}
	\input{Sections/07_experiments.tex}

\input{Sections/08_conclusions.tex}
	\input{Sections/09_acknowledgements.tex}

	\bibliography{pfpe}
	\bibliographystyle{icml2023}
	

	\newpage
	\appendix
	\onecolumn
	\input{Appendix/appendix.tex}

\input{Appendix/appendix_experiments.tex}

\end{document}

%% file: Sections/00_abstract.tex

\begin{abstract}
	
Integral to many recent successes in deep reinforcement learning has been a class of temporal difference methods that use infrequently updated \textit{target values} for policy evaluation in a Markov Decision Process. At the same time, a complete theoretical explanation for the effectiveness of target networks remains elusive. In this work, we provide an analysis of this popular class of algorithms, to finally answer the question:  ``why do target networks stabilise TD learning''? To do so, we formalise the notion of a \textit{partially fitted policy evaluation} method, which describes the use of target networks and bridges the gap between fitted methods and semigradient temporal difference algorithms. Using this framework we are able to uniquely characterise the so-called \textit{deadly triad}--the use of TD updates with (nonlinear) function approximation and off-policy data--which often leads to nonconvergent algorithms. This insight leads us to conclude that the use of target networks can mitigate the effects of poor conditioning in the Jacobian of the TD update. Furthermore, we show that under mild regularity conditions and a well tuned target network update frequency, convergence can be guaranteed even in the extremely challenging off-policy sampling and nonlinear function approximation setting.
\end{abstract}

%% file: Sections/01_introduction.tex
\section{Introduction}
Since their introduction in deep $Q$-networks (DQN) a decade ago \citep{mnih2013atari,mnih2015human}, target networks have become a common feature of state-of-the-art deep reinforcement learning algorithms \citep{lillicrap16ddpg,haarnoja17sql,haarnoja18sac,fujimotoDQL18}. Theoretical analysis of target networks has been limited and there has been no satisfactory explanation for their empirical success in stabilising policy evaluation algorithms.  Whilst recent analysis has characterised the convergence properties of policy evaluation using target networks \citep{lee2019targetbased,fan2020theoretical,zhang2021breaking}, existing approaches focus on asymptotic results, and usually make simplifying assumptions that neither hold in practice nor account for the true behaviour of target network-based updates. Our work finds that the use of target networks can guarantee that deep RL algorithms will not diverge, even in regimes where traditional RL algorithms fail. Additionally, we establish the first finite-time performance bounds for target networks and general function approximation---without strong simplifying assumptions. Moreover, we prove our key stability assumption can always be satisfied by augmenting our updates with simple $\ell_2$ regularisation that does not change the TD fixed points. In doing so, we finally provide theoretical justification for the empirical success that has been observed in challenging, off-policy tasks.

To achieve this, we analyse the use of infrequently updated target value functions by characterising them as a family of  methods that we refer to as \textit{partially fitted policy evaluation} (PFPE). This variant bridges the gap between fitted policy evaluation (FPE) \citep{le2019batch}---which iteratively fit the Bellman backups onto the class of representable function approximators ---and classic temporal difference (TD) algorithms \citep{Sutton88td} by limiting the fitting phase to a fixed number of steps, precisely reflecting the periodically updated target network algorithms as used in practice.

To characterise the performance of PFPE, we express our algorithm--which has traditionally been viewed through the lens of two-timescale analysis--using a single update applied only to the target network parameters. We show that the stability of the algorithm is determined by analysing the eigenvalues of the Jacobian of this update. This formulation allows us to characterise both the limiting (asymptotic) and finite-time (non-asymptotic) convergence properties of PFPE. Furthermore, it suggests, counterintuitively, that target networks are actually the object being optimised rather than merely a means to stabilise conventional TD updates. This insight leads us to empirically investigate a novel target parameter update scheme that uses a momentum-style update \citep{polyak1964momentum}, setting the stage for future research of practical target-based algorithms.

Our bounds on the finite-time performance of PFPE apply to off-policy, nonlinear and partially fitted methods, which have never been investigated previously. We develop key insights into the usefulness of target networks, which we find do not improve asymptotic performance when decaying step sizes are used. Instead, target networks improve the conditioning of TD and fitted methods when the step size \emph{does not tend to zero}, as is often implemented in practice. Under non-decaying stepsizes, our Jacobian analysis shows how PFPE reconditions the TD Jacobian allowing us to prove convergence in regimes where classic TD methods are unstable, thereby breaking the so-called deadly triad that has plagued TD methods \citep{Sutton2018}.  Furthermore, our results do not depend on unwieldy assumptions or modifications of algorithms used in practice, such as projection, bounded state spaces, linear function approximation, or iterate averaging, as is done in previous analysis. In addition to our theoretical results, we experimentally evaluate our bounds on a toy domain, indicating that they are tight under relevant hyperparameter regimes.  Taken together, our results lead to novel insight as to how exactly target networks affect optimisation, and when and why they are effective, leading to actionable results that can be used to further future research.

%% file: Sections/02_preliminaries.tex
\section{Preliminaries}
\label{sec:preliminaries}
\emph{Proofs for all theorems, propositions and corollaries can be found in \cref{app:proofs}}

We denote the set of all probability distributions on a set $\mathcal{X}$ as $\mathcal{P}(\mathcal{X})$. We use $\lVert \cdot \rVert$ to denote the $\ell_2$-norm. For a matrix $M$, we denote the set of eigenvalues as $\lambda(M)$ with the set of maximum normed eigenvalues as $\lambda_{\max}(M)\coloneqq \argsup_{\lambda'\in\lambda(M)}\lvert\lambda'\rvert$ and $\lambda_{\min}(M)\coloneqq \arginf_{\lambda'\in\lambda(M)}\lvert\lambda'\rvert$. The $\ell_2$-norm (spectral norm) for matrix $M$ is $\lVert M \rVert = \sqrt{\lambda_{\max}(M^\top M )}$. Given a function $f:\mathcal{X}\to \mathbb{R}$ and a distribution $\mu\in\mathcal{P}(\mathcal{X})$, we denote the $L_2$-norm as: $\lVert f \rVert_\mu\coloneqq \sqrt{\mathbb{E}_{x\sim\mu}\left[f(x)^2\right]}$.

\subsection{Reinforcement Learning}
We consider the infinite horizon discounted RL setting. The agent interacts with an environment, formalised as a  Markov Decision Process (MDP):	$\mathcal{M}\coloneqq\langle \mathcal{S}, \mathcal{A}, P, P_0, R,\gamma \rangle$ with state space $\mathcal{S}$, action space $\mathcal{A}$, transition kernel $P:\mathcal{S}\times\mathcal{A}\to \mathcal{P}(\mathcal{S})$, initial state distribution $P_0\in\mathcal{P}(\mathcal{S})$, bounded stochastic reward kernel $R:\mathcal{S}\times\mathcal{A} \to \mathcal{P}([-r_{max}, r_{max}])$ where $r_{max}\in\mathbb{R}<\infty$ and scalar discount factor $\gamma \in [0,1)$. An agent in state $s\in\mathcal{S}$ taking action $a\in\mathcal{A}$ observes a reward $r\sim R(s,a)$. The agent's behaviour is determined by a policy that maps a state  to a distribution over actions: $
\pi:\mathcal{S}\to \mathcal{P}(\mathcal{A})$ and the agent transitions to a new state $s'\sim P(s,a)$. We denote the joint distribution of $s',a',r$ conditioned on $s,a$ for policy $\pi$ as $P^\pi_{sar}(s,a)$. We seek to optimise (in the control case), or estimate (in the policy evaluation case) the expected discounted sum of future rewards starting from a given state $s\in\mathcal{S}$. This quantity is given by the state value function, $V^\pi(s) = \mathbb{E}_{a\sim\pi(s)}\left[Q^\pi(s,a)\right]$, with $Q^\pi:\mathcal{S}\times\mathcal{A}\to [-r_{max}/(1 - \gamma), r_{max}/(1 - \gamma)]$, the action value function, given recursively through the Bellman equation: $Q^\pi(s,a) =\mathcal{T}^\pi[Q^\pi](s, a)$,
where the Bellman operator $\mathcal{T}^\pi$ projects functions forwards by one step through the dynamics of the MDP:
\begin{equation}
	\mathcal{T}^\pi[Q^\pi](s, a) \coloneqq\mathbb{E}_{s',a',r\sim P_{sar}^\pi(s,a)}\left[r + \gamma Q^\pi(s',a')\right].
\end{equation} 
$\mathcal{T}^\pi$ is a $\gamma$-contractive mapping and thus has a fixed point, which corresponds to the true value of $\pi$ \citep{puterman2014markov}. When estimating MDP values, we employ a value function approximation $Q_\omega:\mathcal{S}\times\mathcal{A}\to \mathbb{R} $ parametrised by $\omega\in\Omega \subseteq \mathbb{R}^n$.

Many RL algorithms employ TD learning for policy evaluation, which combines bootstrapping, state samples and sampled rewards to estimate the expectation in the Bellman operator \citep{Sutton88td}. In their simplest form, TD methods update the function approximation parameters according to:
\begin{align}
	&\omega_{i+1} = \\
	&\quad\omega_i + \alpha_i \left(r + \gamma Q_{\omega_i}(s',a') - Q_{\omega_i}(s,a)  \right)\nabla_\omega Q_{\omega_i}(s,a),
\end{align}
 where $s\sim d, a\sim \mu(s), s',  a',r \sim P^\pi_{sar}(s,a)$, $d\in\mathcal{P}(\mathcal{S})$ is a sampling distribution, and $\mu$ is a sampling policy that may be different from the target policy $\pi$. For simplicity of notation and to accommodate the introduction of target networks in \cref{sec:pfpe}, we define the tuple $\varsigma\coloneqq(s,a,r,s',a')$ with distribution $P_\varsigma$ and the TD-error vector as:
\begin{align}
		\delta(\omega,\omega',\varsigma)\coloneqq \left(r + \gamma Q_{\omega'}(s',a') - Q_{\omega}(s,a)  \right)\nabla_\omega Q_{\omega}(s,a),
\end{align}
allowing us to write the TD parameter update as:
\begin{align}
	\omega_{i+1} = \omega_i + \alpha_i \delta(\omega_i,\omega_i,\varsigma).
\end{align}
We make the following i.i.d. assumption for clarity of exposition, but discuss other sampling regimes in \cref{app:extensions}:
\begin{assumption}\label{ass:iid}
	Each $s\sim d$ is drawn i.i.d..
\end{assumption}
Typically, $d$ is the steady-state distribution of an ergodic Markov chain. We denote the expected TD-error vector as: $\delta(\omega,\omega')\coloneqq \mathbb{E}_{\varsigma\sim P_\varsigma}[\delta(\omega,\omega',\varsigma)]$
and define the set of TD fixed points as:
\begin{align}
	\omega^\star \in\Omega^\star\coloneqq \left\{ \omega \vert \delta(\omega,\omega)=0\right\}.\label{eq:td_fixed_point}
\end{align}
If a TD algorithm converges, it converges to a TD fixed point. Convergence of TD methods can only be guaranteed for linear function approximators when sampling on-policy in an ergodic MDP, that is the agent sampling and target distributions are the same. We investigate the phenomenon further as part of our asymptotic analysis in \cref{sec:deadly_triad}.
\vspace{-0.7cm}

%% file: Sections/03_pfpe.tex
\section{Partially Fitted Policy Evaluation}
\label{sec:pfpe}
Unfortunately, real-world applications of RL often demand the expressiveness of nonlinear function approximators like neural networks and/or the ability to use data that has been collected off-policy, i.e., by following a policy $\mu$ that differs from the target policy $\pi$ for policy evaluation. 

\subsection{Fitted v Partially Fitted Policy Evaluation}
Fitted methods improve on the sample efficiency and stability of TD methods by explicitly incorporating the limitations of the function approximation class through the use of a projection operator \citep{tsitsiklis97}. These methods generally perform some variant of the iterate $Q_{\bar{\omega}_{l+1}} = \Pi^{d^\pi}\mathcal{T}^\pi Q_{\bar{\omega}_{l}}$ where $\Pi^d$ is the projection operator
$\Pi^{d}Q = \argmin_{Q'} \lVert Q' - Q \rVert_{d,\mu}$. These updates are known as fitted policy evaluation (PFE).

The projection step is needed to accommodate the fact that values generally cannot be exactly represented with function approximation. To obtain a practical way of carrying out the PFE updates, a separate set of \emph{target parameters} can be introduced $\bar{\omega}_l\in\Omega$ that parameterise the TD target and are updated every $k$ timesteps:
\vspace{-0.1cm}
\begin{align}
	\omega_{kl+i+1} &= \omega_{kl+i} + \alpha_{kl+i} \delta(\omega_{kl+i},\bar{\omega}_l,\varsigma),\label{eq:approximator_update}\\
	\bar{\omega}_{l+1}&=
	\omega_{k(l+1)},
	\label{eq:target_update}
\end{align} 
The function approximator update in \cref{eq:approximator_update} carries out $k$ iterations of stochastic gradient descent (SGD) on the loss:
\begin{align}
	\mathcal{L}(\omega;\bar{\omega}_l)\coloneqq \lVert Q_\omega -\mathcal{T}^\pi[Q_{\bar{\omega}_l}] \rVert_{d,\mu},
\end{align}
before updating the target parameters. In the limit as $k\rightarrow \infty$, assuming convergence of SGD to a global minimum, fully fitted policy evaluation occurs by finding $\omega_\infty\in\arginf_{\omega\in\Omega}\mathcal{L}(\omega,\bar{\omega}_l)$.

In practice $k$ is finite and only partial policy evaluation occurs before updating the target parameters, a setting we call partially fitted policy evaluation (PFPE). Without loss of generality, we assume that $\bar{\omega}_0$ is deterministic with $\lVert \bar{\omega}_0\rVert<\infty$ and $\alpha_i=\alpha_l$ for all $kl \le i <k(l+1)$, that is stepsizes only change after updating target parameters. As the target parameters are updated to the approximator parameters every $k$ timesteps in \cref{eq:target_update}, it suffices to consider the target parameter update in isolation when analysing PFPE. Our goal is thus to analyse a single update for the target parameters in the canonical form: 
\begin{align}
	\bar{\omega}_{l+1} =  g^k(\bar{\omega}_l,\mathcal{D},\alpha_l),\quad \mathcal{D}\sim P_\mathcal{D},\label{eq:target_canonical}
\end{align}
where $\mathcal{D}\coloneqq \{ \varsigma_i\}_{i=1}^{k}$ is a set of $k$ samples from the environment with distribution $P_\mathcal{D}$ and $g^k(\bar{\omega}_l,\mathcal{D}_l,\alpha_l)$ reduces the $k$ nested updates from \cref{eq:approximator_update} into a single update for the target parameters.
\subsection{Jacobian Analysis}
\label{sec:Jacobian_analysis}
In our analysis, we show that the stability of the expected PFPE update 
$g^k(\bar{\omega}_l,\alpha_l)\coloneqq \mathbb{E}_{\mathcal{D}\sim P_\mathcal{D}}\left[g^k(\bar{\omega}_l,\mathcal{D},\alpha_l)\right]$ is determined by the conditioning of three Jacobians. We denote the Hessian of the loss as: $	H(\omega;\bar{\omega}_l)\coloneqq \nabla^2_\omega\mathcal{L}(\omega;\bar{\omega}_l)$, the Jacobian of the TD-error vector as: $J_\delta(\omega;\bar{\omega}_l)\coloneqq \nabla_{\omega'}\delta(\omega,\omega')\vert_{\omega'=\bar{\omega}_l}$ and define the TD Jacobian as: $J_\textrm{TD}(\bar{\omega}_l)\coloneqq \nabla_{\omega} \delta(\omega,\omega)\vert_{\omega=\bar{\omega}_l}$. Observe that $J_\textrm{TD}(\bar{\omega}_l)= J_\delta(\bar{\omega}_l,\bar{\omega}_l)-H(\bar{\omega}_l;\bar{\omega}_l)$. Without loss of generality, we assume that the Hessian matrix is diagonalisable because, if it is not, an arbitrarily small perturbation can make its eigenvalues distinct and therefore diagonalisable.  So that these matrices exist, we require that the expected PFPE update is differentiable almost everywhere, a condition that is guaranteed by a Lipschitz assumption. We also require that the variance of the updates is bounded, motivating the following regularity assumption:
\begin{assumption} [Function Approximator Regularity]\label{ass:regularity} We assume that $\delta(\omega,\omega',\varsigma)$ is Lipschitz in $\omega,\omega'$ with constant $L$:
	$\lVert \delta(\omega_1,\omega'_1,\varsigma)-\delta(\omega_2,\omega'_2,\varsigma) \rVert\le L(\lVert\omega_1-\omega_2 \rVert +\lVert\omega_1'-\omega_2' \rVert )$
	 and  $\Omega$ is convex, $\mathbb{V}_{\varsigma\sim P_\varsigma}[\delta(\omega,\omega,\varsigma)]\coloneqq\mathbb{E}_{\varsigma\sim P_\varsigma}[\lvert\delta(\omega,\omega,\varsigma)-\delta(\omega,\omega)\rVert^2]\le\sigma^2_\delta $ for some $\sigma_\delta^2<\infty$.
\end{assumption}
The bounded variance assumption can easily be achieved for unbounded function approximators by truncating the TD error vector, much like the commonly used gradient clipping in gradient descent. We now introduce the path-mean Jacobians, which are the principal element of our analysis:
\begin{align}
		\bar{H}(\omega,\omega^\star;\bar{\omega}_l)&\coloneqq- \int^1_0 \nabla_{\omega'} \delta(\omega'=\omega-t(\omega-\omega^\star),\bar{\omega}_l)dt,\\
	\bar{J}_\delta(\omega,\omega^\star;\bar{\omega}_l)&\coloneqq  \int_0^1\nabla_{\omega'}\delta(\bar{\omega}_l,\omega'=\omega-t(\omega-\omega^\star)) dt,\\	\bar{J}_\textrm{TD}(\omega,\omega^\star)&\coloneqq 	\int_0^1\nabla_{\omega'}\delta(\omega',\omega')\vert_{\omega'=\omega-t(\omega-\omega^\star)} dt.
\end{align}
Intuitively, a path-mean Jacobian is the average of all of the Jacobians along the line joining $\omega$ to $\omega^\star$. The convexity assumption in \cref{ass:regularity} ensures that the line integral joining any two points in $\Omega$ always exists.  The Lipschitz assumption in \cref{ass:regularity}  is only required for \cref{sec:asymptotic_pfpe} and can be weakened to any condition that ensures the path-mean Jacobians exist for the remainder of the paper. 

Our analysis in \cref{sec:asymptotic_pfpe} proves that stability of TD and PFPE under decaying stepsizes is determined solely by the negative definiteness of the TD path-mean Jacobian $\bar{J}_\textrm{TD}(\omega,\omega^\star)$.  In \cref{sec:non-asymptotic}, we show for a non-diminishing stepsize regime that through suitable regularisation (which does not affect the TD fixed point), PFPE's stability can be determined \emph{only} by $\alpha_l$ and $k$, for which stable values exists. As $\bar{H}(\omega,\omega^\star;\bar{\omega}_l)$ is the path-mean Hessian of the loss, convergence can be guaranteed under the same mild assumptions required to prove convergence of a stochastic gradient descent algorithm to minimise $\mathcal{L}(\omega;\bar{\omega}_l)$. This implies that PFPE can converge under regimes where TD will not as $\bar{J}_\textrm{TD}(\omega,\omega^\star)$ is positive definite. 
\subsection{Analysis of PFE}
\label{sec:pfe_analysis}
 We now showcase the power of our Jacobian analysis by writing the PFE updates exactly in terms of $(\bar{\omega}_0-\omega^\star)$:
\begin{theorem}\label{proof:PFE_stabilty} Under \cref{ass:regularity}, the sequence of PFE updates $\bar{\omega}_{l+1}^\star \in \arginf_{\omega} \mathcal{L}(\omega,\bar{\omega}_l^\star)$ satisfy:
	\begin{align}
		 &\bar{\omega}_l^\star-\omega^\star\\
		 &\quad= \prod_{i=0}^{l-1}\left( \bar{H}(\bar{\omega}_{i+1}^\star,\omega^\star;\bar{\omega}_i^\star)^{-1} \bar{J}_\delta(\bar{\omega}_i^\star,\omega^\star;\omega^\star)\right)(\bar{\omega}_0-\omega^\star).
	\end{align}
\end{theorem}
We can use \cref{proof:PFE_stabilty} to determine the stability of FPE updates. If $\sup_{\omega,\omega'\in\Omega}\left\lVert\bar{H}(\omega',\omega^\star;\omega)^{-1} \bar{J}_\delta(\omega,\omega^\star;\omega^\star)\right\rVert<1$ then the FPE updates are a contraction mapping and will converge to a fixed point under the Banach fixed-point theorem. We discuss the convergence of FPE under varying regularisation schemes in \cref{sec:stabilising_FPE}.

%% file: Sections/04_asymptotic_analysis.tex
\section{Asymptotic Analysis}
\label{sec:asymptotic_pfpe}
We now study the behaviour of \cref{eq:target_canonical} in the limit of $l\rightarrow \infty$.  We introduce the standard Robbins-Munro condition for the decaying stepsizes that is a necessary condition to ensure convergence to a fixed point:
\begin{assumption}[Robbins-Munro]\label{ass:stepsizes}
	Each $\alpha_l$ is a positive scalar with $\sum_{l=0}^\infty \alpha_l =\infty$ and $\sum_{l=0}^\infty \alpha_l^2 <\infty$. 
\end{assumption} 
Now we introduce a core necessary assumption to prove stability of PFPE with diminishing stepsizes:
\begin{assumption}[TD Stability]\label{ass:TD_jacobian_stability} There exists a region $\mathcal{X}_{\textrm{TD}}(\omega^\star)$ containing a fixed point $\omega^\star$  such that  
	$\bar{J}_\textrm{TD}(\omega,\omega^\star)$ has strictly negative eigenvalues for all $\omega\in\mathcal{X}_{\textrm{TD}}(\omega^\star)$.
\end{assumption}
The key insight from \cref{ass:TD_jacobian_stability} is that the stability of PFPE under diminishing stepsizes is determined only by the eigenvalues of the single step path-mean Jacobian $\bar{J}_\textrm{TD}(\omega,\omega^\star)$, regardless of the value of $k$ or $\alpha_l$. Indeed, stochastic approximation can be shown to be provably divergent if this condition cannot be satisfied \citep{Pemantle90}. From this perspective, if TD diverges then so will 
PFPE under \emph{diminishing stepsizes}, hence the asymptotic stability of PFPE is independent of $k$ and $\alpha_l$, and, unlike updating under a two-timescale regime, introducing target parameters that are updated periodically every $k$ timesteps does not improve asymptotic convergence properties under this analysis. Once \cref{ass:TD_jacobian_stability} has been established, there are several approaches to prove convergence of the PFPE update under varying sampling conditions and projection assumptions. We follow the proof of \citep{vidyasagar22}, but discuss approaches that generalise our assumptions in \cref{app:extensions}
\begin{theorem}\label{proof:asymptotic}
 Let Assumptions~\ref{ass:iid} to~\ref{ass:TD_jacobian_stability} hold. If there exists some fixed point $\omega^\star$ with region of contraction $\mathcal{X}_{\textrm{TD}}(\omega^\star)$ and timestep $t$ such that $\bar{\omega}_l \in \mathcal{X}_{\textrm{TD}}(\omega^\star)$ for all $l\ge t$ the  the sequence of target parameter updates in \cref{eq:target_update} converge almost surely to $\omega^\star$. 
\end{theorem}
\subsection{The Deadly Triad}
\label{sec:deadly_triad}
We have established that it is not possible to prove  convergence of PFPE under diminishing stepsizes if \cref{ass:TD_jacobian_stability} does not hold. We now discuss how adherence to \cref{ass:TD_jacobian_stability} formalises  a phenomenon known as the \emph{deadly triad} \citep{Sutton2018} where it has been established that TD cannot be proved to converge when using function approximators in the off-policy setting. To control for the effect of nonlinear function approximation, we first investigate linear function approximators of the form $Q_\omega(s,a)=\phi(s,a)^\top \omega$ where $\phi:\mathcal{S}\times\mathcal{A}\rightarrow \mathbb{R}^n$ is a feature vector. Define the one-step lookahead distribution as: $P^\mu \coloneqq\mathbb{E}_{s\sim d,a\sim \mu(s)} \left[P(s,a) \right]$. Introducing the shorthand:
\begin{align}
	\Phi&\coloneqq \mathbb{E}_{s\sim d, a\sim\mu(s)}[\phi(s,a)\phi(s,a)^\top],\\
	\Phi'&\coloneqq \mathbb{E}_{s\sim d, a\sim\mu(s)}[\mathbb{E}_{s'\sim P^\mu, a'\sim\pi(s')}[\phi(s',a')]\phi(s,a)^\top],
\end{align}
we can derive the TD Jacobian as:
\begin{align}
\bar{J}_\textrm{TD}(\omega,\omega^\star)=\gamma\Phi'-\Phi.
\end{align}
We now examine why the conditioning of  $	\bar{J}_\textrm{TD}(\omega,\omega^\star)$ explains this phenomenon.
\paragraph {Linear Function Approximation}
\label{sec:linear_triad}
For linear function approximators, we show in \cref{app:derivation_equivalent} that $\gamma	\lVert Q_\omega \rVert_{P^\mu,\pi}< \lVert Q_\omega \rVert_{d,\mu}$ for all $\omega$ is a sufficient condition for  $\gamma\Phi'-\Phi$ to have negative eigenvalues, thereby satisfying \cref{ass:TD_jacobian_stability}. This implies that the function approximator class remains non-expansive under the one-step lookahead distribution $P^\mu$, thereby preventing the function approximator diverging as the Markov chain is traversed. This condition has been introduced previously in the fitted $Q$-iteration literature \citep{wang2020statistical, wang2021instabilities} as a ``low distribution shift'' assumption. 

In the on-policy setting in an ergodic MDP, we can prove that there exists a stationary distribution $d^\pi$ induced by following the target policy $\pi$, that is $\mu=\pi$. Moreover  it is assumed that samples come from $d^\pi$; hence by the definition of ergodicity, the one-step lookahead distribution is the stationary distribution: $P^\pi=d^\pi$. It thus follows that $\gamma\lVert Q_\omega \rVert_{P^\mu,\pi}=\gamma\lVert Q_\omega \rVert_{ d^\pi,\pi}<\lVert Q_\omega \rVert_{ d^\pi,\pi}$ and hence  \cref{ass:TD_jacobian_stability} holds automatically for on-policy TD in an ergodic MDP, thereby establishing the convergence properties as a special case via \cref{proof:asymptotic}. 

For off-policy data, it is not possible to prove that  $\gamma	\lVert Q_\omega \rVert_{P^\mu,\pi}< \lVert Q_\omega \rVert_{d,\mu}$ holds without further assumptions on the sampling policy and MDP. In general, it is not possible to show that $\bar{J}_\textrm{TD}(\omega,\omega^\star)$ is negative definite in the off-policy case as the distribution shift may be too high: there exist  counterexample MDPs where off-policy algorithms such as $Q$-learning provably diverge under linear function approximation \citep{Williams1993Analysis,Baird1995}. 
\paragraph{Nonlinear Function Approximation}
Even in an on-policy regime, we cannot prove convergence of TD when nonlinear function approximators such as neural networks are used. In these cases, the path-mean Jacobian may not have a closed form solution. However, it can be bounded by the following norm (see \cref{sec:nonlinear_Jacobian_derivation}):
	\begin{align}
	&\sup\lambda\left(\bar{J}_\textrm{TD}(\omega,\omega^\star)\right)\\
	&\qquad \le\sup_\omega\sup\lambda \left(
			\mathbb{E}\left[
				(\mathcal{T}^\pi[Q_\omega]-Q_\omega )
				\nabla^2_\omega Q_{\omega}
			\right]
		 \right. \\ &\qquad\qquad\qquad \left.
			+\mathbb{E}\left[
				\left(
					\gamma\mathbb{E}'[\nabla_\omega Q_\omega']
					-\nabla_\omega Q_\omega
				\right)
				\nabla_\omega Q_\omega^\top
			\right]
	\right).
\end{align}
Even making the same assumption as in \cref{sec:linear_triad} of sampling on-policy in an ergodic MDP to show that
\begin{align}
&\omega^\top\mathbb{E}\left[\left(\gamma\mathbb{E}'[\nabla_\omega Q_\omega'] -\nabla_\omega Q_\omega\right)\nabla_\omega Q_\omega^\top
\right]\omega\\
&\qquad\le (\gamma-1) \omega^\top \mathbb{E}\left[ \nabla_\omega Q_\omega \nabla_\omega Q_\omega^\top\right]\omega<0,
\end{align}
 we cannot prove the negative definiteness of $\bar{J}_\textrm{TD}(\omega,\omega^\star)$  required to satisfy \cref{ass:TD_jacobian_stability}. This is because the matrix $\mathbb{E}\left[(\mathcal{T}^\pi[Q_\omega]-Q_\omega )\nabla^2_\omega Q_{\omega}\right]$ can be arbitrarily positive definite depending on the MDP and choice of function approximator. Indeed, there exist counterexample MDPs with provably divergent nonlinear function approximators when sampling on-policy \citep{tsitsiklis97}.

%% file: Sections/05_finite_analysis.tex
\section{Non-asymptotic Analysis}
\label{sec:non-asymptotic}
Our asymptotic analysis in \cref{sec:asymptotic_pfpe} shows that increasing $k$ or adjusting $\alpha_l$ for PFPE does not affect the asymptotic strong convergence properties of the TD algorithm, implying that target networks do not stabilise TD if stepsizes tend to zero. We showed that the underlying reason for this was the deadly triad, which we formalised as adherence to \cref{ass:TD_jacobian_stability}. We now replace \cref{ass:TD_jacobian_stability}, that is $\bar{J}_\textrm{TD}(\omega,\omega^\star)$ is negative definite, with the assumption that FPE is stable: 

\begin{assumption}[FPE Stability]\label{ass:FPE_stability} There exists a region $\mathcal{X}_{\textrm{FPE}}(\omega^\star)$ containing a fixed point $\omega^\star$  such that  
	$\sup_{\omega,\omega'\in\mathcal{X}_{\textrm{FPE}}(\omega^\star)}\left\lVert\bar{H}(\omega',\omega^\star;\omega)^{-1} \bar{J}_\delta(\omega,\omega^\star;\omega^\star)\right\rVert<1$.
\end{assumption}

\subsection{Stabilising FPE}
\label{sec:stabilising_FPE}
We now prove that \cref{ass:FPE_stability} can always be satisfied using regularisation schemes that do not affect the TD fixed points. We introduce the following regularised TD vector:
\begin{align}
		\delta_\textrm{Reg}(\omega,\omega') =\delta(\omega,\omega')+\rho(\omega,\omega'),\label{eq:td_regularised}
\end{align}
where $\rho(\omega,\omega')$ is a regularisation term such that $\rho(\omega,\omega)=0$, thereby not changing the TD fixed point or TD update. As an example, $\rho(\omega',\omega')$ can contain powers of regularisation terms $M_\textrm{Reg}(\omega-\omega')$ in addition to combinations of $\delta(\omega',\omega)$  and $\delta(\omega,\omega')$ terms, where $\delta(\omega',\omega)$ is a TD vector with target and $Q$-network parameters swapped. In this paper, we briefly study regularisation of the form:
\begin{align}
	\delta_\textrm{Reg}(\omega,\omega')=&\mu \delta(\omega,\omega') \\&+(1-\mu)\left(\delta(\omega',\omega)-\eta(\omega-\omega')\right), \label{eq:regularised_update}
\end{align}
where $\mu$ mixes the TD updates and $\eta$ controls the degree of regularisation. We emphasise that $\delta_\textrm{Reg}(\bar{\omega}_l,\bar{\omega}_l)=\delta(\bar{\omega}_l,\bar{\omega}_l)$, leaving the TD update unchanged. In contrast, unless $\omega^\star$ is known a priori,  introducing regularisation that modifies the TD update---as is done in \citep{zhang2021breaking}---will affect the TD fixed points. We now prove that FPE can be stabilised by treating $\eta$ and $\mu$ as hyperparameters to be tuned to the specific MDP. 
\begin{proposition}\label{proof:stabilising_FPE}
Using the regularised TD vector in \cref{eq:regularised_update}, the path-mean Jacobians are:
\begin{align}
		\bar{H}_\textrm{Reg}(\omega,\omega^\star;\bar{\omega}_l) &=	\mu\bar{H}(\omega,\omega^\star;\bar{\omega}_l)\\
		&\quad-(1-\mu) \left(\bar{J}_{\delta}(\omega,\omega^\star;\bar{\omega}_l)- I\eta\right),\\
\bar{J}_{\delta,\textrm{Reg}}(\omega,\omega^\star;\bar{\omega}_l)&=\mu \bar{J}_{\delta}(\omega,\omega^\star;\bar{\omega}_l)\\
&\quad-(1-\mu)\left(\bar{H}(\omega,\omega^\star;\bar{\omega}_l)- I\eta\right).
\end{align}
\cref{ass:FPE_stability} is satisfied if:
\begin{align}
	\sup_{\omega,\omega'\in \mathcal{X}_{\textrm{FPE}}(\omega^\star)}\left\lVert  	\bar{H}_\textrm{Reg}(\omega',\omega^\star;\omega) ^{-1}	\bar{J}_{\delta,\textrm{Reg}}(\omega,\omega^\star;\omega^\star)\right\rVert < 1.\label{eq:regularised_condition}
\end{align}
There exists finite $\eta,\mu$ such that \cref{eq:regularised_condition} holds.
\end{proposition}
The key insight from \cref{proof:stabilising_FPE} is that regularisation stabilises FPE (and hence PFPE) without affecting existing TD fixed points, even when TD is unstable, motivating future research directions to develop sophisticated regularisation techniques.

\subsection{Convergence Analysis}
By carrying out a non-asymptotic analysis, we now investigate how the deadly triad can be broken by PFPE using \cref{eq:td_regularised} when stepsizes \emph{do not tend to zero}. This leads to a formal understanding of how target parameters stabilise TD under stepsize regimes that are actually used in practice when classic TD methods fail. The foundation of our analysis is a condition function that can be used to determine the stability of the updates:

\begin{definition}[Condition Function]\label{def:condition_function} For a subset $\mathcal{X}(\omega^\star) \subseteq \Omega$ with corresponding fixed point $\omega^\star\in\mathcal{X}(\omega^\star)$ such that $\omega_i\in \mathcal{X}(\omega^\star)$ for all $i\ge0$ , let 
	\vspace{-0.1cm}
	\begin{align}
		\lambda_H^\star&\coloneqq \sup_{\omega,\omega',\omega''} \argsup_{ \lambda'\in\lambda(\bar{H}(\omega,\omega';\omega''))}\left\lvert 1-\alpha_l \lambda'\right\rvert,\\
		\left\lVert\bar{J}_\textrm{FPE}^\star\right\rVert&\coloneqq \sup_{\omega,\omega'\in\mathcal{X}(\omega^\star)} \left\lVert\bar{H}(\omega',\omega^\star;\omega)^{-1} \bar{J}_\delta(\omega,\omega^\star;\omega')\right\rVert ,\\
		\left\lVert\bar{J}_\textrm{TD}^\star\right\rVert&\coloneqq 	\sup_{\omega\in\mathcal{X}(\omega^\star)}\lVert I+\alpha\bar{J}_\textrm{TD}(\omega,\omega^\star)\rVert,
	\end{align}
	and define the condition function as:
	\begin{align}
		\mathcal{C}(\alpha_l,k)
		&\coloneqq 	\left\lvert 1-\alpha_l \lambda_H^\star	\right\rvert^{k-1}\left\lVert\bar{J}_\textrm{TD}^\star\right\rVert\\
		&\quad+\left(1+	\left\lvert 1-\alpha_l \lambda_H^\star	\right\rvert^{k-1}\right)\left\lVert \bar{J}_\textrm{FPE}^\star\right\rVert.\label{eq:condition_function}
	\end{align}
\end{definition}
The condition function depends on the maximal eigenvectors of the Jacobians introduced in \cref{sec:Jacobian_analysis}, and so can still be used to analyse general nonlinear function approximators for which the path-mean Jacobians have no analytic solution. Using the condition function, we decompose the error at a given timestep into the effect of the expected update plus the error induced by variance of the update:
\begin{theorem}
	Define 
\begin{align}
	\sigma_k&\coloneqq
	\left(1-\left\lvert 1-\alpha_l \lambda_H^\star	\right\rvert^k\right)\frac{\sigma_\delta}{\lambda_H^\star},
\end{align}
Let Assumptions~\ref{ass:iid} and~\ref{ass:regularity} hold, then:
\begin{align}
	\mathbb{E}\left[\lVert\bar{\omega}_{l+1}-\omega^\star\rVert \right]\le \mathcal{C}(\alpha_l,k)\mathbb{E}\left[\lVert\bar{\omega}_l-\omega^\star\rVert\right]+\alpha_l\sigma_k.\label{eq:expected_difference_equivalent}
\end{align}
\end{theorem}

 The effect of the expected update (the first term in \cref{eq:expected_difference_equivalent}) is bounded by the condition function, which depends both on data conditioning but critically, on both $k$ and $\alpha_l$ as well and must diminish with increasing $l$ to ensure convergence. Using this decomposition, we see convergence is guaranteed if the following assumption holds:

\begin{assumption}[Contraction Region]\label{ass:contraction_region}
	We assume that $\mathcal{C}(\alpha,k)\le c<1$ over $\mathcal{X}_\textrm{FPE}(\omega^\star)$.
\end{assumption}
allowing us to prove convergence of PFPE for stepsizes that don't tend to zero provided that updates remain in a region of contraction:
\begin{corollary} \label{proof:nonasy_convergence} Let Assumptions~\ref{ass:iid},~\ref{ass:regularity},~\ref{ass:FPE_stability} and~\ref{ass:contraction_region} hold.  For a fixed stepsize  $\alpha_l=\alpha>0$,
	\begin{align}
		\mathbb{E}&\left[
		\lVert \bar{\omega}_l- \omega^\star\rVert
		\right]
		\le \frac{\alpha\sigma_k}{1-c}\\
		&\quad\quad+\exp(-l(1-c))\left(
		\left\lVert
		\bar{\omega}_0- \omega^\star
		\right\rVert
		-\frac{\sigma_k}{1-c}
		\right)	.
	\end{align}
\end{corollary}
\cref{proof:nonasy_convergence} is a key result of this work. Our result demonstrates geometric decay of errors in $l$, to a ball of fixed radius $\frac{\alpha\sigma_k}{1-c}$. This is analogous to related work in stochastic gradient descent \citep{bottou2018optimization}, and matches the intuition that, without decaying stepsize, variance in the updates means that convergence to a fixed point does not occur. Note that the radius of the ball which we converge to can be made arbitrarily small by decreasing $\alpha$. 

This supports the use of a hybrid approach, wherein a fixed step size is used until iterates are no longer improving and then reducing step size and repeating to decrease the radius of the ball of convergence whilst maintaining $k$ as small as possible. In the remainder of this section, we explore the properties of the condition function to ensure the existence of a region of contraction satisfying \cref{ass:contraction_region}.  
\subsection{Properties of PFPE Condition Function}
\label{sec:nonlinear_nonasymptotic}
We now investigate key properties of \cref{eq:condition_function} to understand how target parameters can lead to convergence when classic TD methods fail. If $\bar{J}_\textrm{TD}(\omega,\omega^\star)$ is positive definite, TD is provably divergent, however our analysis reveals that there are values of $k$  and  $\alpha_l$ for which PFPE does converge.

\paragraph{Property 1: Lower bound}		$\left\lVert\bar{J}_\textrm{FPE}^\star\right\rVert\le \mathcal{C}(\alpha_l,k)$.

We first investigate the conditions for which our choice of function approximators can never be used to prove convergence. Our condition function implies that we cannot prove convergence for any $\lambda_H^\star\le0$ or $\lambda_H^\star\ge \frac{2}{\alpha_l}$ as repeated applications of $\lvert 1-\alpha_l\lambda_H^\star\rvert^2$ do not reduce the effect the ill-conditioning of $\bar{J}_\textrm{TD}(\omega,\omega^\star)$. We formalise this in the following regularity assumption: 
\begin{assumption}[Eigenvalue Regularity Assumption]\label{ass:eigenvalues}
	Given a region $\mathcal{X}\subseteq\Omega$, for all $\omega,\omega'\in\mathcal{X}$ there exists $0<\lambda_1^{\min}$ and $\lambda_1^{\max}<\infty$ such that $\lambda^{\min}\le\lambda(\nabla_\omega^2 \mathcal{L}(\omega;\omega'))\le\lambda^{\max}$.
\end{assumption}
We now propose two simple fixes to avoid this issue.  Recall from \cref{sec:Jacobian_analysis} that $\lambda_H^\star$ is an eigenvalue of the Hessian of a loss. If $\lambda_H^\star$ was negative, this would imply that the Hessian is not positive semidefinite for all $\omega$ in the region of interest; hence we cannot prove convergence of stochastic gradient descent on the loss $\mathcal{L}(\omega;\bar{\omega}_l)$, let alone the full PFPE algorithm. To remedy this problem, the eigenvalues of the matrix can be increased using the regularisation introduced in \cref{eq:td_regularised} without affecting the TD fixed point.  However, if $\lambda_H^\star\ge \frac{2}{\alpha_l}$, then the conditioning of the Hessian matrix is ill-suited to the chosen step-size, and an easy remedy is to decrease $\alpha_l$. Our bound shows that the condition function is lower bounded by $\lVert J_\textrm{FPE}^\star\rVert$, and so if \cref{ass:FPE_stability} does not hold, then convergence of PFPE is not provable.

 \paragraph{Property 2: Monotonicity} 
 
For $|1-\alpha_l\lambda_H^\star\rvert<1$, $\mathcal{C}(\alpha_l,k)\le \mathcal{C}(\alpha_l,k')$ for $k\le k'$. 

The monotonicity property ensures that $|1-\alpha_l\lambda_H^\star\rvert<1$ defines the interval of Hessian eigenvalues for which there is a regime in which we can increase $k$ in order to ensure PFPE updates are a contraction mapping. This suggests that a key role of the target network is to help mitigate the effects of the ill-conditioning of the TD Jacobian when using fixed step sizes. We now investigate how decreasing stepsizes and increasing the number of PFPE steps affect the conditioning of PFPE,
which validates this hypothesis.

\paragraph{Property 3: Limits}
For any $k<\infty$, $\lim_{\alpha_l\rightarrow 0}\mathcal{C}(\alpha_l,k)=\left\lVert\bar{J}_\textrm{TD}^\star\right\rVert+2\left\lVert \bar{J}_\textrm{FPE}^\star\right\rVert$. For any $0<\alpha_l<\frac{2}{\lambda_H^\star}$, $\lim_{k\rightarrow \infty} \mathcal{C}(\alpha_l,k)=\left\lVert\bar{J}_\textrm{FPE}^\star\right\rVert$.

The first limit illustrates the effects of a diminishing stepsize sequence, confirming our bound is consistent with the results of the previous section that increasing $k$ does not improve the convergence properties of PFPE if stepsizes tend to zero and PFPE only stabilises TD for $0<\alpha_l$. By taking the limit $k\rightarrow\infty$, we compliment our monotonicity result, obtaining a bound for how much we can improve on the stability of TD by increasing $k$.  As expected, in the limit of $k\rightarrow\infty$, the condition function tends to $\lVert J_\textrm{FPE}^\star\rVert$. Through this insight, we interpret PFPE as mixing FPE and TD updates according the coefficient $\lvert 1-\alpha_l \lambda^\star_H\rvert^{k-1}$: for $k=1$, PFPE uses only TD updates and in the limit $k\rightarrow\infty$, PFPE recovers the FPE update.

\subsection{Breaking the Deadly Triad}
\label{sec:breaking_the_triad}

\begin{figure}[h]
	\centering
	\includegraphics[width=\linewidth]{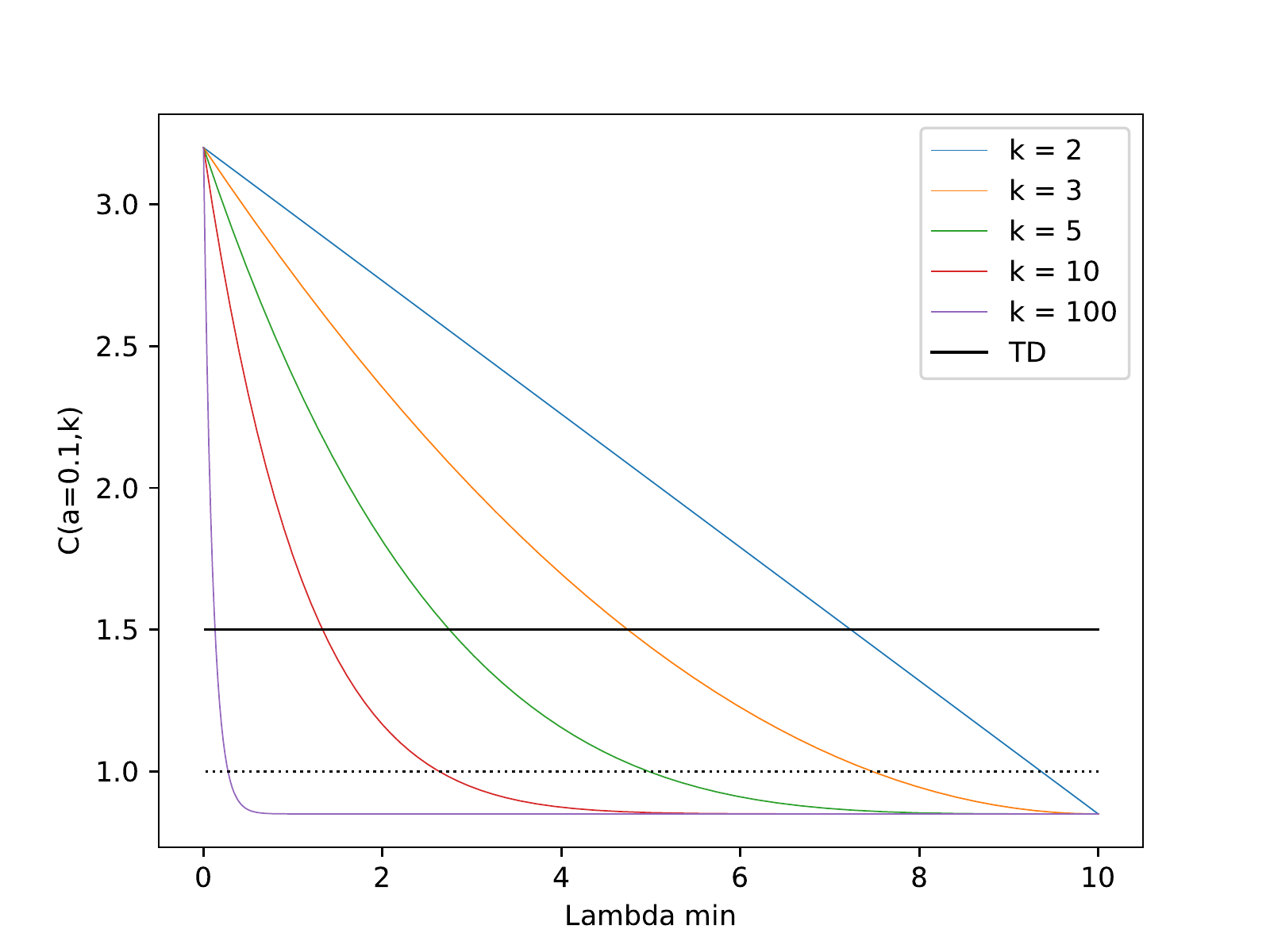}
	\caption{ We plot $\mathcal{C}(\alpha=0.1,k)$ for $\lVert \bar{J}^\star_\textrm{FPE} \rVert=0.85$  and  $\lVert \bar{J}_\textrm{TD}^\star\rVert\le 1.5$  with increasing $k$ as a function of $\lambda_{\min}$.  }
	\label{fig:convergence}
\end{figure}

We now combine all properties presented in this section into our main result, proving that through suitable regularisation and choice of $\alpha_l$ and $k$, PFPE breaks TD's deadly triad described in \cref{sec:deadly_triad}:
\begin{theorem} \label{proof:convergent_stepizes} Let \cref{ass:eigenvalues} hold over $\mathcal{X}_\textrm{FPE}(\omega^\star)$ from \cref{def:condition_function}. 
For any $\frac{1}{\alpha_l}>\frac{\lambda_1^{\min}+\lambda_1^{\max}}{2}$ such that $\alpha_l>0$, any
\begin{align}
	k>1+ \frac{\log(1-\lVert\bar{J}_\textrm{FPE}^\star\rVert)-\log(\lVert \bar{J}_\textrm{TD}^\star\rVert+\lVert\bar{J}_\textrm{FPE}^\star\rVert) }{\log(1-\alpha \lambda^\textrm{min})},
\end{align}
ensures that $\mathcal{X}_\textrm{FPE}(\omega^\star)$ is a region of contraction satisfying \cref{ass:contraction_region}.
\end{theorem}
\cref{proof:convergent_stepizes} demonstrates that appropriate values of $\alpha_l$ and $k$ can be found by treating them as hyperparameters, decreasing  $\alpha_l$  and increasing $k$ until the algorithm is stable, reducing the conditions needed to prove convergence of PFPE to those of proving convergence of stochastic gradient descent on the loss $\mathcal{L}(\omega;\bar{\omega}_l)$. The key insight of \cref{proof:convergent_stepizes} is that even when TD is unstable due to $1<\lVert I +\alpha_l \bar{J}_\textrm{TD}(\bar{\omega}_l,\omega^\star) \rVert$, there exists a finite $	k$ such that $ \mathcal{C}(\alpha_l,k)<1$ and hence PFPE is stable. We illustrate this phenomenon with a sketch in \cref{fig:convergence}, demonstrating that increasing $k$ ensures PFPE is provably convergent in regimes where TD cannot be proved to converge. 

The key insight of our analysis is that, unlike in TD where stability can only be proved if the matrix  $\bar{J}_\delta(\omega,\omega^\star;\omega^\star)-\bar{H}(\omega,\omega^\star;\omega)$ is negative definite, with suitable regularisation, the stability of PFPE can be determined solely by tuning $\alpha_l$ and $k$, regardless of the MDP, sampling regime, or function approximator, thereby breaking the deadly triad. The choice of $\alpha_l$ and $k$ thus becomes a trade-off between maintaining a fast rate of convergence and reducing the residual variance $(\alpha_l\sigma_k)^2$ in \cref{eq:expected_difference_equivalent}.

%% file: Sections/06_related_work.tex
\section{Related Work}
\label{sec:related_work}
Our work furthers the analysis of TD, FPE, and target-network based methods. In this section we provide a brief overview of previous investigations of these algorithms. 
\paragraph{Fitted Policy Evaluation}
FPE is a relatively well understood class of RL algorithms from a theoretical perspective. \citet{nedic2003least} analyse the convergence of the Least-Squares Policy Evaluation (LSPE) of \citet{bertsekas1996temporal} in an on-policy, linear function approximation setting. Analysis of LSPE shows that learning with constant step size leads to theoretical and empirical gains compared to TD and LSPE with decaying step sizes \citep{bertsekas2004improved}, which mirrors our conclusions in \cref{sec:breaking_the_triad}.

In the context of fitted methods applied to off-policy and control problems, \citet{munos2008finite} prove generalisation properties of Fitted $Q$ Iteration \citep{ernst2005tree} for general function classes under assumptions of low projection error and limited data distribution shift. \citet{le2019batch} coin the term FPE, and formalise the algorithm for general function approximators, with theoretical results under similar assumptions to \citet{munos2008finite}.
 \vspace{-0.5cm}
\paragraph{Theory of TD}
Previous results concerning convergence rates of classic TD methods largely argue that the Bellman operator is a contraction, and thus most focus on linear function approximation. \citet{tsitsiklis97} first proved convergence of linear, on-policy TD, arguing that the projected Bellman operator in this setting is a contraction. This corresponds to a special case of \cref{ass:TD_jacobian_stability}. \citet{dalal2017finite} give the first finite time bounds for linear TD(0), under an i.i.d.\ data model similar to the one that we use here. \citet{bh2018finite} provide bounds for linear TD in both the i.i.d.\ data setting and a correlated data setting, through analogy with SGD. \citet{srikant2019finite} approach the problem from the perspective of Ordinary Differential Equations (ODE) analysis, bounding the divergence of a Lyapunov function from the limiting point of the ODE that arises from the TD update scheme.
 \vspace{-0.3cm}
\paragraph{Analysis of Target Networks}
Existing analysis of the theoretical properties of target networks are limited, usually involving algorithmic changes or restrictive assumptions. \citet{yang2019convergent} show convergence of a $Q$-learning approach using a target network that is updated using Polyak averaging with nonlinear function approximation. However their analysis--which makes use of two-timescale analysis--requires a projection step to limit the magnitude of parameters. \citet{carvalho2020convergent} show convergence of a related method using two-timescale analysis, though their target network update differs significantly from those used in practice. \citet{zhang2021breaking} analyse the use of target networks with linear function approximation, but require projection steps on both the target network and value parameters. \citet{lee2019targetbased} provide finite-iteration bounds, but are limited to on-policy data, linear function approximation, and near-perfect fitting to the target network between updates.
\citet{fan2020theoretical} analyse the use of target networks for deep Q-learning \citep{mnih2015human} with the simplifying assumption that they are performing some form of Fitted $Q$ Iteration.

None of these efforts yield finite time bounds with target networks, nor do any match the policy evaluation methods used in practice as well as the PFPE analysis studied here. Furthermore, our use of a single target network update, rather than independent target and value updates leads to simpler bounds without the need for a two-timescale analysis.
\paragraph{GTD and TDC Methods} While not directly related to PFPE or the use of target networks, GTD-style approaches \citep{sutton2008convergent,sutton2009fast,maei2009convergent} also lead to convergent, TD-style algorithms, even with off-policy sampling or nonlinear function approximation. These methods maintain a second set of parameters which must be optimised at a faster timescale than the value parameters. However, these approaches are commonly found to be ineffective and not used in practice due to the difficulty in tuning the rate of second timescale (see, e.g. \citet{fellows2021bayesian}), and potentially additional variance introduced by the second set of parameters \citep{ghiassian2020gradient}.

\paragraph{Improving Conditioning of TD Methods}
Previous work concerning conditioning of TD methods has been largely concerned with approximation of preconditioning approaches to iterative-methods \citep{saad2003iterative}. The first such approach was focused on preconditioning of on-policy, linear, least-squares forms of TD \citep{yao2008preconditioned}. \citet{chen2020zap,romoff2020tdprop} adapt this approach for nonlinear function approximation, though their results are still on-policy. Our work, on the other hand, demonstrates that use of the target network, alongside fixed step sizes, changes the form of parameter iterates to ameliorate the poor conditioning that occurs when directly applying TD or fitted methods, even in off-policy settings.

%% file: Sections/07_experiments.tex
\section{Experiments}
\label{sec:experiments}

We proceed to empirical investigation of our bounds. First, we demonstrate that the use of an infrequently updated target network leads to convergence of off-policy evaluation on the Baird's notorious counterexample. Then, we evaluate the effect of a speculative modified update rule in the Cartpole-v0 ``gym'' environment \citep{brockman2016openai}. Additional implementation details for both experiments can be found in Appendix \ref{app:experiments}.

\subsection{Baird's Counterexample}
In this experiment, we demonstrate the practicality of our core claim--that for sufficiently high $k$ and low enough $\alpha$, PFPE will not diverge, even under conditions that TD does. To do so, we evaluate the use of target networks with varying update frequencies on the well known off-policy counterexample due to \citet{baird1995residual}.

In this environment, depicted in \cref{app:experiments}, rewards are zero everywhere, transitions are deterministic, and the true solution lies within the linear function approximation class that we make use of. The behaviour policy is set such that all states are sampled with uniform probability. The target policy, however, always transitions to a specific state, and remains there. Due to undersampling of this absorbing state, conventional TD policy evaluation diverges, demonstrating that even in simple environments, TD can be unstable when applied off policy with function approximation.

\begin{figure}[h]
\centering
\includegraphics[width=0.7\linewidth]{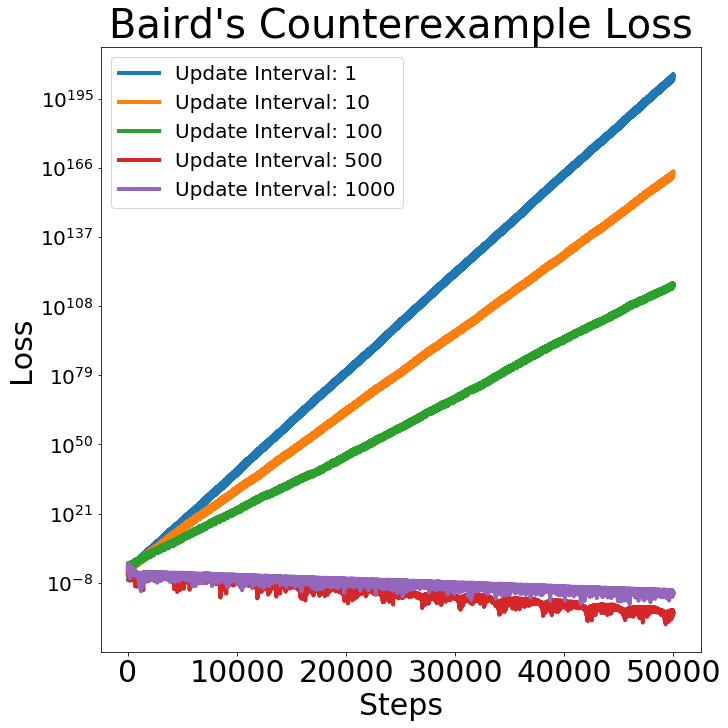}
\caption{Experiment on Baird's counterexample. Decreasing the frequency of target network updates improves conditioning and leads to convergence of PFPE for suitable choices of hyperparameters.}
\label{fig:baird_loss}
\end{figure}

We report the stepwise (fitted) error in \cref{fig:baird_loss} across different values of $k$, for fixed step size $\alpha=0.01$, and fixed discount factor $\gamma=0.99$. We see that with $k=1$--which is equivalent to using TD with fixed step sizes--our parameters diverge. Likewise, if $k$ is set to 5 or 10, we are unable to overcome the conditioning of the TD Jacobian and diverge, albeit at a slower rate. Once we take $k \geq 500$, however, conditioning has improved enough to lead to convergence. This supports our theoretical conclusion: that PFPE can be used to improve the convergence conditions of TD.

\subsection{Cartpole Experiment}

One important insight of our analysis is that we can view the entire optimisation process as a sequence of updates to the target network only. This suggests investigation into alternative forms or acceleration of target network updates. Inspired by the use of optimisation methods with momentum in RL settings \citep{sarigul2018performance, haarnoja18sac}, we investigate the effects of a target network that is updated using momentum.

Unlike the standard periodic target network update in \cref{eq:target_update}, we postulate that there may be settings in which a periodic update with momentum may accelerate or stabilise convergence. This update works as follows:
\begin{align}
    \bar{\omega}&=\begin{cases}
        (1 - \mu) \omega_i + \mu (\omega_{i-k} - \omega_{i-2k} ),& i\bmod k=0,\\
        \bar{\omega}, & \textrm{otherwise}.
    \end{cases}
    \label{eq:momentum_target_update}
\end{align}
\begin{figure}[h]
	\centering
	\includegraphics[width=\linewidth]{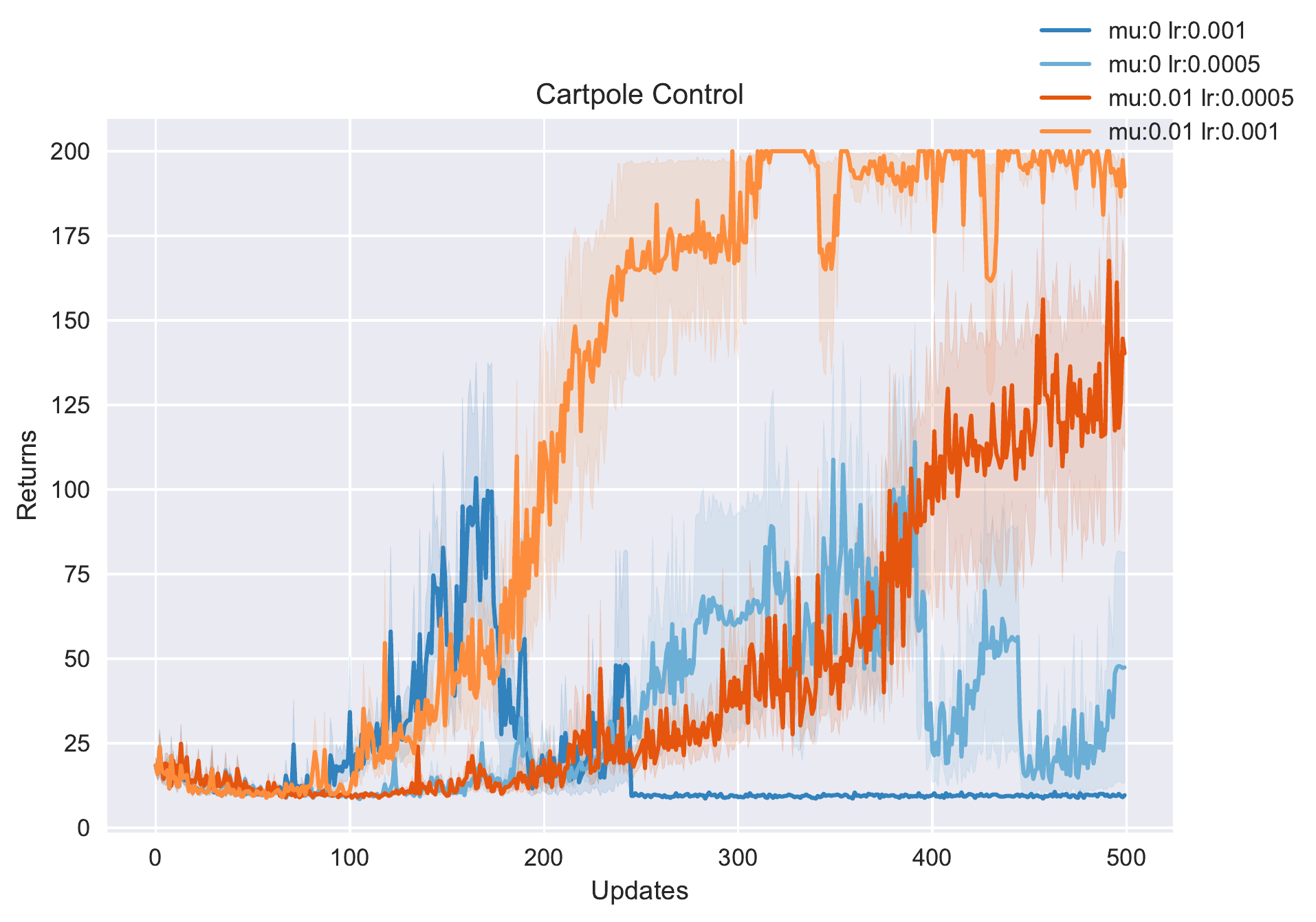}
	\caption{Cartpole Experiment. The agent with the momentum update is significantly more stable and able to consistently learn, while without the modified update, learning collapses.}
	\label{fig:cartpole}
\end{figure}
We investigate the effects of this momentum update on the Cartpole domain. For this experiment, we use control results in which the policy is continuously learned. This is because control problems are inherently off-policy, and induce additional instability, and thus benefit from faster and more stable convergence of values. We implement the standard DQN \citep{mnih2015human} algorithm, with our modified target network update in order to examine its effect. The results are shown in Figure \ref{fig:cartpole}. Our proposed update indeed leads to improved learning and stability, at least for the hyperparameter ranges tested, suggesting that the momentum update has merit. As a result, we propose investigation of more sophisticated target network update schemes as an avenue for future research.

%% file: Sections/08_conclusions.tex
\section{Conclusions}
\label{sec:conclusions}
This work analysed the use of target networks through the formulation of a novel class of TD updates, which we refer to as PFPE. These updates generalise traditional TD(0) and fitted policy evaluation methods. Our analysis contributes asymptotic and finite time bounds without additional restrictive assumptions or significant changes to the algorithms used in practice. In our main result, we uncovered novel insight as to when and how target networks are useful: provided step-sizes don't tend to zero and FPE is stable, there always exists a finite number of update steps $k$ and non-zero upper bound over stepsizes such that PFPE can improve conditioning to ensure learning is stable when classic TD methods fail. Our focus on the target network update as the object of concern in terms of optimisation suggests that novel, accelerated methods for updating target networks may help speed up and stabilise learning. Our initial experiments support this notion. Moreover, our analysis reveals that regularisation may be key to determining the stability of PFPE, opening a promising avenue for future research.  

%% file: Sections/09_acknowledgements.tex
\section*{Acknowledgements}
Mattie Fellows is funded by a generous grant from Waymo. We would like to thank Valentin Thomas for providing a helpful discussion.

%% file: Appendix/appendix.tex
\appendix

\setcounter{theorem}{0}
\setcounter{proposition}{0}
\setcounter{lemma}{0}

\section{Derivations}

\subsection{Derivation of \cref{ass:TD_jacobian_stability} from low distributional shift}

\label{app:derivation_equivalent}
Starting from \cref{ass:TD_jacobian_stability} and the definition of negative definiteness, we need to show:
\begin{align}
	\omega^\top(\gamma\Phi'-\Phi)\omega < 0,
\end{align}
whenever $\gamma \lVert Q_\omega \rVert_{P^\mu,\pi}< \lVert Q_\omega \rVert_{d,\mu}$, for all $\omega$. Investigating the first term by expanding the expectations we see:
\begin{align}
	\gamma\omega^\top\Phi'\omega &= \gamma\mathbb{E}_{s\sim d, a\sim\pi(s)}\left[
	\omega^\top \phi(s, a)
	\mathbb{E}_{s'\sim P(s, a), a'\sim\pi(s')}\left[
	\phi(s', a')^\top\omega
	\right]
	\right], \\
	&= \gamma \mathbb{E}_{d, \pi, P^\mu}\left[\omega^\top\phi(s, a) \phi(s', a')^\top\omega\right],\\
	&\leq \gamma\sqrt{\mathbb{E}_{d, \pi, P^\mu}\left[(\phi(s, a)^\top\omega)^2\right]\mathbb{E}_{d, \pi, P^\mu}\left[(\phi(s', a')^\top\omega)^2\right]},\\
	&\leq \gamma\sqrt{\mathbb{E}_{d, \pi}\left[(\phi(s, a)^\top\omega)^2\right]\mathbb{E}_{d, \pi, P^\mu}\left[(\phi(s', a')^\top\omega)^2\right]}, \\
	&\leq \gamma  \lVert Q_\omega \rVert_{d}\lVert Q_\omega \rVert_{P^\mu,\pi}.
\end{align}
This allows us to apply our assumption:
\begin{align}
	\omega^\top(\gamma\Phi'-\Phi)\omega  \leq \gamma \lVert Q_\omega \rVert_{P^\mu,\pi} \lVert Q_\omega \rVert_{d} -
	\lVert Q_\omega \rVert_{d}^2 \le  \gamma\lVert Q_\omega \rVert_{d}^2 - \lVert Q_\omega \rVert_{d,\mu}^2 < 0.
\end{align}

\subsection{Nonlinear Jacobian Analysis}
\label{sec:nonlinear_Jacobian_derivation}
We start by bounding the maximum eigenvalue:
\begin{align}
\sup\lambda\left(\bar{J}_\textrm{TD}(\omega,\omega^\star)\right)	&=\sup_{\omega}\frac{\omega^\top\bar{J}_\textrm{TD}(\omega,\omega^\star)\omega}{ \omega^\top\omega},\\
&=\sup_\omega\int_0^1  \frac{\omega^\top J_\textrm{TD} (\omega'-t(\omega'-\omega^\star))\omega}{\omega \omega^\top}  dt,\\
&\le\int_0^1 \sup_\omega \frac{\omega^\top J_\textrm{TD} (\omega'-t(\omega'-\omega^\star))\omega}{\omega \omega^\top}  dt,\\
&\le\int_0^1\sup_{t\in[0,1]} \sup_\omega \frac{\omega^\top J_\textrm{TD} (\omega'-t(\omega'-\omega^\star))\omega}{\omega \omega^\top}  dt,\\
&=\sup_{t\in[0,1]} \sup_\omega \frac{\omega^\top J_\textrm{TD} (\omega'-t(\omega'-\omega^\star))\omega}{\omega \omega^\top}  \underbrace{\int_0^1 dt}_{=1},\\
&\le\sup_{\omega'} \sup_\omega \frac{\omega^\top J_\textrm{TD} (\omega'-t(\omega'-\omega^\star))\omega}{\omega \omega^\top}   ,\\
&=\sup_\omega\sup\lambda\left(J_\textrm{TD}(\omega,\omega^\star)\right).
\end{align}
 We now substitute for the definition of the TD Jacobian, yielding:
\begin{align}
J_\textrm{TD}(\omega,\omega^\star)&=	  \nabla_{\omega} \delta(\omega,\omega),\\
	=	&\nabla_{\omega} 	\mathbb{E}_{\varsigma\sim P_\varsigma}\left[\left(r + \gamma Q_{\omega}(s',a') - Q_{\omega}(s,a)  \right)\nabla_\omega Q_{\omega}(s,a)\right],\\
=&	\mathbb{E}_{\varsigma\sim P_\varsigma}\left[\left( \gamma\nabla_{\omega} Q_{\omega}(s',a') - \nabla_{\omega}Q_{\omega}(s,a)  \right)\nabla_\omega Q_{\omega}(s,a)+\left(r + \gamma Q_{\omega}(s',a') - Q_{\omega}(s,a)  \right)\nabla_\omega^2 Q_{\omega}(s,a)\right],\\
=&	\mathbb{E}_{\varsigma\sim P_\varsigma}\left[\left( \gamma\nabla_{\omega} Q_{\omega}(s',a') - \nabla_{\omega}Q_{\omega}(s,a)  \right)\nabla_\omega Q_{\omega}(s,a)+\left((\mathcal{T}^\pi[Q_\omega](s,a) - Q_{\omega}(s,a)  \right)\nabla_\omega^2 Q_{\omega}(s,a)\right],
\end{align}
as required.

\section{Proofs}
\label{app:proofs}

\subsection{FPE Analysis}
\label{app:projection_operator_conditioning}

\begin{lemma}\label{proof_app:pfe_one_step} Under \cref{ass:regularity}, the FPE update  $\bar{\omega}_{l+1} \in \arginf_{\omega} \mathcal{L}(\omega,\bar{\omega}_l)$ satisfies:
	\begin{align}
		\bar{\omega}_l^\star-\omega^\star=  \bar{H}(\bar{\omega}_{l}^\star,\omega^\star;\bar{\omega}_l)^{-1} \bar{J}_\delta(\bar{\omega}_l,\omega^\star;\omega^\star)(\bar{\omega}_l-\omega^\star),\label{eq:FPE_update_exact}
	\end{align}
	\begin{proof}
		Given $\bar{\omega}_{l}$, the FPE fixed point $\bar{\omega}_l^\star$ must be an element of the set:
		\begin{align}
			\bar{\omega}_l^\star\in \{\omega \vert \delta(\omega,\bar{\omega}_{l})=0 \},
		\end{align}
		which we use to derive a stability condition for the projection operator:
		\begin{align}
			\delta(\bar{\omega}_l^\star,\bar{\omega}_{l})&=\delta(\omega^\star,\omega^\star)=0\\
			\implies\delta(\bar{\omega}_l^\star,\bar{\omega}_l)-\delta(\omega^\star,\bar{\omega}_{l})&=\delta(\omega^\star,\omega^\star)-\delta(\omega^\star,\bar{\omega}_{l}).
		\end{align}
		Let $\ell_1(t)\coloneqq \bar{\omega}_l^\star-t(\bar{\omega}_l^\star-\omega^\star)$ and $\ell_2(t)\coloneqq \bar{\omega}_{l}-t(\bar{\omega}_{l}-\omega^\star)$. We introduce the notation:
		\begin{align}
			\delta_1(t,\bar{\omega}_{l})\coloneqq \delta(\ell_1(t),\bar{\omega}_{l}),\quad\delta_2(t,\omega^\star)\coloneqq \delta(\omega^\star,\ell_2(t)).
		\end{align}
		We observe that $	\delta_1(0,\bar{\omega}_{l})=\delta(\bar{\omega}_l^\star,\bar{\omega}_{l})$ and $	\delta_1(1,\bar{\omega}_{l})=\delta(\omega^\star,\bar{\omega}_{l})$, and $	\delta_2(0,\omega^\star)=\delta(\omega^\star,\bar{\omega}_{l})$ and $	\delta_2(1,\omega^\star)=\delta(\omega^\star,\omega^\star)$. From the fundamental theorem of calculus and \cref{ass:regularity}, it follows:  
		\begin{gather}
			\delta_1(0,\bar{\omega}_{l})-\delta_1(1,\bar{\omega}_{l})=\delta_2(1,\omega^\star)-\delta_2(0,\omega^\star),\\
			\implies	-\int_0^1 \partial_t \delta(\omega=\ell_1(t),\bar{\omega}_{l}) dt= \int_0^1 \partial_t \delta(\omega^\star,\omega=\ell_2(t) )dt,\\
			\implies	\int_0^1 \nabla_\omega \delta(\omega=\ell_1(t),\bar{\omega}_{l}) (\bar{\omega}_l^\star-\omega^\star)dt= -\int_0^1 \nabla_\omega \delta(\omega^\star,\omega=\ell_2(t) )(\bar{\omega}_{l}-\omega^\star)dt,\\
			\implies	-\int_0^1 \nabla_\omega^2 \mathcal{L}(\omega=\ell_1(t);\bar{\omega}_{l}) (\bar{\omega}_l^\star-\omega^\star))dt= -\int_0^1 \nabla_\omega \delta(\omega^\star,\omega=\ell_2(t) )(\bar{\omega}_{l}-\omega^\star)dt,\\
			\implies	\int_0^1 \nabla_\omega^2 \mathcal{L}(\omega=\ell_1(t);\bar{\omega}_{l}) dt (\bar{\omega}_l^\star-\omega^\star)= \int_0^1 \nabla_\omega \delta(\omega^\star,\omega=\ell_2(t) )dt(\bar{\omega}_{l}-\omega^\star),\\
			\implies \bar{H}(\bar{\omega}_l^\star,\omega^\star;\bar{\omega}_{l}) (\bar{\omega}_l^\star-\omega^\star)= 	\bar{J}_\delta(\bar{\omega}_l,\omega^\star;\omega^\star)(\bar{\omega}_{l}-\omega^\star),\\
			\implies	 (\bar{\omega}_l^\star-\omega^\star)=  \bar{H}(\bar{\omega}_l^\star,\omega^\star;\bar{\omega}_{l})^{-1} \bar{J}_\delta(\bar{\omega}_l,\omega^\star;\omega^\star)(\bar{\omega}_l-\omega^\star),
		\end{gather}
		as required.
	\end{proof}
\end{lemma}
\begin{theorem}\label{proof_app:pfe_stability} Under \cref{ass:regularity}, the sequence of FPE updates $\bar{\omega}_{l+1}^\star \in \arginf_{\omega} \mathcal{L}(\omega,\bar{\omega}_l^\star)$ satisfy:
\begin{align}
	\bar{\omega}_l^\star-\omega^\star= \prod_{i=0}^{l-1}\left( \bar{H}(\bar{\omega}_{i+1}^\star,\omega^\star;\bar{\omega}_i^\star)^{-1} \bar{J}_\delta(\bar{\omega}_i^\star,\omega^\star;\omega^\star)\right)(\bar{\omega}_0-\omega^\star).
\end{align}
\begin{proof}
From  \cref{eq:FPE_update_exact}  of \cref{proof_app:pfe_one_step}, it follows:
\begin{align}
	\bar{\omega}_{i+1}^\star-\omega^\star=\bar{H}(\bar{\omega}_{i+1}^\star,\omega^\star;\bar{\omega}^\star_i)^{-1} \bar{J}_\delta(\bar{\omega}^\star_i,\omega^\star;\omega^\star)(\bar{\omega}^\star_i-\omega^\star).
\end{align}
Recursively applying the result $l$ times, our result follows immediately. 
\end{proof}
\end{theorem}

\subsection{Asymptotic Analysis}
\label{app:asymptotic}
For this section, we define a Martingale difference sequence that captures the behaviour of our updates. 	Let $\{\omega_i\}_{i=0}^k$ denote the intermediate function approximation parameters between target parameter updates $\bar{\omega}_{l+1}$ and $\bar{\omega}_l$, with $\omega_0=\bar{\omega}_l$ and $\omega_k=\bar{\omega}_{l+1}$. We start by writing our target parameter updates as: 
\begin{align}
	\omega_1&=\bar{\omega}_l+\alpha_l\delta(\bar{\omega}_l,\bar{\omega}_l,\varsigma_0),\\
	\omega_2&=\omega_1+\alpha_l\delta(\omega_1,\bar{\omega}_l,\varsigma_1),\\
	&=\bar{\omega}_l+\alpha_l\left(\delta(\bar{\omega}_l,\bar{\omega}_l,\varsigma_0)+\delta(\bar{\omega}_l+\alpha_l\delta(\bar{\omega}_l,\bar{\omega}_l,\varsigma_0),\bar{\omega}_l,\varsigma_1)\right),\\
	\omega_3&=\omega_2+\alpha_l\delta(\omega_2,\bar{\omega}_l,\varsigma_2),\\
	&=\bar{\omega}_l+\alpha_l\left(\delta(\bar{\omega}_l,\bar{\omega}_l,\varsigma_0)+\delta(\bar{\omega}_l+\alpha_l\delta(\bar{\omega}_l,\bar{\omega}_l,\varsigma_0),\bar{\omega}_l,\varsigma_1)\right)\\
	&\qquad+\alpha_l(\delta(\bar{\omega}_l+\alpha_l\left(\delta(\bar{\omega}_l,\bar{\omega}_l,\varsigma_0)+\delta(\bar{\omega}_l+\alpha_l\delta(\bar{\omega}_l,\bar{\omega}_l,\varsigma_0),\bar{\omega}_l,\varsigma_1)\right),\bar{\omega}_l,\varsigma_2),\\
	&\ \  \vdots\\
	\omega_k&=\bar{\omega}_l+\alpha_l\sum_{i=0}^{k-1}\delta(\bar{\omega}_l+\alpha_l h_i(\bar{\omega}_l,\mathcal{D},\alpha_l) ,\bar{\omega}_l,\varsigma_i),\\
	&=\bar{\omega}_l+\alpha_l h_k(\bar{\omega}_l,\mathcal{D},\alpha_l),
\end{align}
where we define $h_i(\bar{\omega}_l,\mathcal{D},\alpha_l)$ recursively as:
\begin{align}
	h_i(\bar{\omega}_l,\mathcal{D},\alpha_l)\coloneqq \sum_{j=0}^{i-1}\delta(\bar{\omega}_l+\alpha_l h_j(\bar{\omega}_l,\mathcal{D},\alpha_l),\bar{\omega}_l,\varsigma_j ).
\end{align}
and remark that $h_0(\bar{\omega}_l,\mathcal{D},\alpha_l)=0$ trivially. We write our target parameters updates as:
\begin{align}
	\bar{\omega}_{l+1} =\omega_k=  \bar{\omega}_{l}  +\alpha_l \left(k \delta(\bar{\omega}_l,\bar{\omega}_l) + \mathcal{M}_{l+1} + \varepsilon_{l+1}\right),\label{eq:nonlinearupdate_app}
\end{align}
where 
\begin{align}
	\varepsilon_{l+1}\coloneqq 	h_k(\bar{\omega}_l,\mathcal{D}_l,\alpha_l)- \sum_{i=0}^{k-1} \delta(\bar{ \omega}_l,\bar{ \omega}_l,\varsigma_i),
\end{align}
and $\mathcal{M}_{l+1}$ defines the Martingale sequence:
\begin{align}
	\mathcal{M}_{l+1}\coloneqq \sum_{i=0}^{k-1} \delta(\bar{ \omega}_l,\bar{ \omega}_l,\varsigma_i)-k \delta(\bar{\omega}_l,\bar{\omega}_l)\label{eq:martingale_defo}
\end{align}
In this section, we demonstrate that the proof of \citet[Theorem 2.2]{borkar00samethod} can be adapted to account for the additional term $\varepsilon_{l+1}$ that arises due to the use of target networks in the updates.  \cref{proof:diminishing_term}   demonstrates that as stepsizes tend to zero, the effect of $\epsilon_{l+1}$ becomes negligible, hence the inclusion of $\varepsilon_{l+1}$ negligible to our analysis of the underlying ODE defined by the TD updates. 
\begin{lemma}\label{proof:diminishing_term}
	Let $\nu_{n,n+m}\coloneqq \sum_{l=n}^{m+n-1}\alpha_l\epsilon_{l+1}$ for $m\ge 1$. Under Assumptions~\ref{ass:iid} to ~\ref{ass:stepsizes},  $\lim_{n\rightarrow\infty}\sup_{m}\lVert \nu_{n,n+m}\rVert=0 $ almost surely. 
	\begin{proof}
		We start by bounding each $\lVert\epsilon_{i+1}\rVert$ using the the Lipschitzness of $\delta$ from \cref{ass:regularity}:
		\begin{align}
			\lVert\epsilon_{l+1}\rVert&=\left \lVert \sum_{i=0}^{k-1}\left(\delta(\bar{\omega}_l+\alpha_l h_i(\bar{\omega}_l,\mathcal{D},\alpha_l),\bar{\omega}_l,\varsigma_i )- \delta(\bar{ \omega}_l,\bar{ \omega}_l,\varsigma_i)\right)\right\rVert,\\
			&\le\sum_{i=0}^{k-1}\left\lVert\delta(\bar{\omega}_l+\alpha_l h_i(\bar{\omega}_l,\mathcal{D},\alpha_l),\bar{\omega}_l,\varsigma_i )- \delta(\bar{ \omega}_l,\bar{ \omega}_l,\varsigma_i)\right\rVert,\\
			&\le\sum_{i=0}^{k-1}L\left\lVert\bar{\omega}_l+\alpha_l h_i(\bar{\omega}_l,\mathcal{D},\alpha_l)-\bar{ \omega}_l\right\rVert,\\
			&=\alpha_lL\sum_{i=0}^{k-1}\left\lVert h_i(\bar{\omega}_l,\mathcal{D},\alpha_l)\right\rVert,
		\end{align}
		To proceed, we recognise that each $\lVert h_i(\bar{\omega}_l,\mathcal{D}_l,\alpha_l)\rVert\le c_h<\infty$ almost surely where $c_h$ is a finite positive constant - otherwise:
		\begin{gather}
			P(\lVert h_i(\bar{\omega}_l,\mathcal{D}_l,\alpha_l)\rVert=\infty) >0\implies \mathbb{E}[\lVert h_i(\bar{\omega}_l,\mathcal{D}_l,\alpha_l)\rVert ]=\infty\implies\mathbb{E}[\lVert h_i(\bar{\omega}_l,\mathcal{D}_l,\alpha_l)\rVert^2 ]=\infty\\
			\implies \mathbb{E}[\lVert \delta(\bar{\omega}_l+\alpha_l h_{j}(\bar{\omega},\mathcal{D},\alpha_l),\bar{\omega}_l,\varsigma_j ) \rVert^2 ]=\infty,\label{eq:bounded_norm_as}
		\end{gather}
		for at least one $i>j$, hence $\mathbb{V}_{\varsigma\sim P_\varsigma}[\delta(\omega,\omega',\varsigma)]=\infty $ for some $\omega,\omega'$ thereby violating \cref{ass:regularity}. Using $c_h$, we bound $\lVert\epsilon_{l+1}\rVert$:
		\begin{align}
			\lVert\epsilon_{l+1}\rVert&\le \alpha_l L  \sum_{i=0}^{k-1} c_h=\alpha_l c_h k L,
		\end{align}
		almost surely. We use this result to bound $\lVert \nu_{n,n+m}\rVert$:
		\begin{align}
			\lVert \nu_{n,n+m}\rVert\le  \sum_{l=n}^{m+n-1}\alpha_l\lVert\epsilon_{l+1}\rVert\le   c_hkL \sum_{l=n}^{m+n-1}{\alpha_l}^2. \label{eq:nu_bound}
		\end{align}
		Now, under \cref{ass:stepsizes}, 
		\begin{align}
			\lim_{n\rightarrow\infty} \sup_{m}\sum_{l=n}^{m+n-1}{\alpha_l}^2=0,
		\end{align}
		hence by the bound established in \cref{eq:nu_bound}:
		\begin{align}
			\lim_{n\rightarrow\infty} \sup_{m}\lVert \nu_{n,n+m}\rVert=0,
		\end{align}
		almost surely, as required.
	\end{proof}
\end{lemma}
\begin{theorem}\label{proof_app:linear_asymptotic}
	Under Assumptions~\ref{ass:iid}-~\ref{ass:TD_jacobian_stability}, the sequence of target parameter updates in \cref{eq:target_update} converge almost surely to $\omega^\star$.
	\begin{proof} 
		Our update
		\begin{align}
			\bar{\omega}_{l+1} =  \bar{\omega}_{l}  +\alpha_l \left(k \delta(\bar{\omega}_l,\bar{\omega}_l) + \mathcal{M}_{l+1} + \varepsilon_{l+1}\right),
		\end{align}
	is identical to the update presented in \citet[Eq. 2.1.1]{borkar00samethod} with an additional term $\varepsilon_{l+1}$. Proof of convergence to the ODE is given by \citet[Lemma 1]{borkar00samethod}, which is predicated on the convergence of:
	\begin{align}
		\Delta_{n,n+m}\coloneqq \zeta_{n+m}-\zeta_n,
	\end{align}
 from \citet[Eq. 2.1.6]{borkar00samethod} where 
		\begin{align}
		\zeta_{n}=\sum_{l=0}^{n-1}\alpha_l\mathcal{M}_{l+1},
	\end{align}
for $n\ge1$, that is $\lim_{n\rightarrow\infty} \sup_{m}\lVert \Delta_{n,n+m}\rVert=0$, almost surely. To adapt our updates so that \citet[Lemma 1]{borkar00samethod} still applies, we recognise that the term $\zeta_{n}$ is now replaced in our updates with: 
\begin{align}
	\bar{\zeta}_{n}=\sum_{l=0}^{n-1}\alpha_l(\mathcal{M}_{l+1}+\epsilon_{l+1}),
\end{align}
and hence $	\Delta_{n,n+m}$ is replaced in our updates with:
	\begin{align}
	\bar{\Delta}_{n,n+m}\coloneqq &\bar{\zeta}_{n+m}-\bar{\zeta}_n,\\
	=& \zeta_{n+m}-\zeta_n + \left(\sum_{l=0}^{n+m-1}\alpha_l\epsilon_{l+1}\right) - \left(\sum_{l=0}^{n-1}\alpha_l\epsilon_{l+1}\right),\\
	=& \zeta_{n+m}-\zeta_n + \sum_{l=n}^{n+m-1}\alpha_l\epsilon_{l+1},\\
	=& \zeta_{n+m}-\zeta_n + \nu_{n,n+m},\\
	=& \Delta_{n,n+m} + \nu_{n,n+m},
\end{align}
where $\nu_{n,n+m}$ is defined as \cref{proof:diminishing_term}. All arguments of \citet[Lemma 1]{borkar00samethod} remain unchanged, except Eq. 2.1.9, where we must now show that $\lim_{n\rightarrow\infty} \sup_{m}\lVert \bar{\Delta}_{n,n+m}\rVert=0$:
\begin{align}
	\lim_{n\rightarrow\infty} \sup_{m}\lVert \bar{\Delta}_{n,n+m}\rVert&\le \lim_{n\rightarrow\infty} \sup_{m}\left(\lVert \Delta_{n,n+m}\rVert + \lVert \nu_{n,n+m}\rVert\right),\\
	&\le \lim_{n\rightarrow\infty} \left(\sup_{m}\lVert \Delta_{n,n+m}\rVert + \sup_{m}\lVert \nu_{n,n+m}\rVert\right),\\
	&= \lim_{n\rightarrow\infty} \sup_{m}\lVert \Delta_{n,n+m}\rVert + \lim_{n\rightarrow\infty}  \sup_{m}\lVert \nu_{n,n+m}\rVert.
\end{align}
Applying \cref{proof:diminishing_term} yields $\lim_{n\rightarrow\infty}  \sup_{m}\lVert \nu_{n,n+m}\rVert=0$ almost surely, hence
\begin{align}
	\lim_{n\rightarrow\infty} \sup_{m}\lVert \bar{\Delta}_{n,n+m}\rVert&\le \lim_{n\rightarrow\infty} \sup_{m}\lVert \Delta_{n,n+m}\rVert,
\end{align}
which is proved in \citet[Lemma 1]{borkar00samethod}. Convergence of our algorithm is thus only predicated on the convergence of the update: 
		\begin{align}
			\bar{\omega}_{l+1} = \bar{\omega}_{l} +\alpha_l \left(k \delta(\bar{\omega}_l,\bar{\omega}_l) + \mathcal{M}_{l+1} \right).\label{eq:alternative_updates}
		\end{align}
 \citet[Theorem 2.2]{borkar00samethod} proves convergence of \cref{eq:alternative_updates} almost surely to $\omega^\star$ given the following four conditions hold:
		\begin{enumerate}[label=\Roman*]
			\item $k\delta (\omega,\omega)$ is Lipschitz in $\omega$,
			\item Stepsizes $\alpha_l$ satisfy \cref{ass:stepsizes},
			\item The sequence $\{\mathcal{M}_l, \mathcal{F}_l\}_{l\ge0}$ is a Martingale difference sequence with respect to the increasing family of $\sigma$-algebras: $\mathcal{F}_l\coloneqq \sigma(\{\bar{\omega}_i,\mathcal{M}_i\}_{i\in\{0:l\}})$ where $\mathbb{E}\left[ \mathcal{M}_{l+1}\vert \mathcal{F}_l \right]=0$ and $\mathbb{E}\left[\lVert \mathcal{M}_{l+1}\rVert^2\vert \mathcal{F}_l \right]\le C (1+\lVert \bar{\omega}_l\rVert^2)$ for some positive $C<\infty$.
			\item The sequence of iterates remain bounded, that is $\sup_l \lVert \bar{\omega}_l\rVert <\infty$ almost surely. 
		\end{enumerate}
		Conditions I and II hold trivially. 
		
		For Condition III, we can take expectations of the Martingale difference:
		\begin{align}
			\mathbb{E}\left[\mathcal{M}_{l+1}\vert \mathcal{F}_l\right]&=\mathbb{E}\left[\mathcal{M}_{l+1}\vert \mathcal{F}_l\right],\\
			&=\mathbb{E}\left[\sum_{i=0}^{k-1} \delta(\bar{ \omega}_l,\bar{ \omega}_l,\varsigma_i)- k \delta(\bar{ \omega}_l)\bigg\vert \mathcal{F}_l\right],\\
			&=\mathbb{E}\left[k \delta(\bar{ \omega}_l,\bar{ \omega}_l)- k \delta(\bar{ \omega}_l,\bar{ \omega}_l)\bigg\vert \mathcal{F}_l\right],\\
			&=0,
		\end{align}
		as required. We now show that the variance is bounded using \cref{ass:regularity}:
		\begin{align}
			\lVert \mathcal{M}_{l+1} \rVert^2&= \left\lVert \sum_{i=0}^{k-1}\left( \delta(\bar{ \omega}_l,\bar{ \omega}_l,\varsigma_i)-  \delta(\bar{ \omega}_l,\bar{\omega}_l)\right)\right\rVert^2,\\
			&\le k\left\lVert \delta(\bar{ \omega}_l,\bar{ \omega}_l,\varsigma_i)-  \delta(\bar{ \omega}_l,\bar{ \omega}_l)\right\rVert^2,\\
		\implies \mathbb{E}\left[\lVert \mathcal{M}_{l+1} \rVert^2\vert \mathcal{F}_l\right]&\le k^2 \mathbb{E}\left[\left\lVert \delta(\bar{ \omega}_l,\bar{ \omega}_l,\varsigma_i)-  \delta(\bar{ \omega}_l,\bar{ \omega}_l)\right\rVert^2\Big\vert \mathcal{F}_l\right],\\
		&=k\mathbb{V}_{\varsigma\sim P_\varsigma}[\delta(\bar{\omega}_l,\bar{\omega}_l,\varsigma)],\\
		&\le k\sigma^2_\delta,
		\end{align}
		thereby satisfying Condition III.
		
		Finally, we prove Condition IV using \citet[Theorem 5]{vidyasagar22}, which states iterates remain bounded almost surely if:
		\begin{enumerate}[label=(\alph*)]
			\item Conditions I and III hold;
			\item there exists some Lyapunov function $V:\Omega \mapsto \mathbb{R}^+$ such that $a\lVert \omega-\omega^\star\rVert^2\le V(\omega)\le b\lVert \omega-\omega^\star\rVert^2$ for constants $a,b>0$ and $\lVert \nabla^2_\omega V(\omega)\rVert$ is bounded, and;
			\item $\nabla_\omega V(\omega)^\top \delta(\omega,\omega)<0$ for all $\omega\in\mathcal{X}_{\textrm{TD}}(\omega^\star)$.
		\end{enumerate}
		  We propose $V(\omega)=\frac{1}{2}\lVert \omega -\omega^\star\rVert^2$ as a candidate Lyapunov function, which trivially satisfies (b). We now show (c) holds by applying the fundamental theorem of calculus to $\delta(\omega,\omega)$. 	Let $\ell(t)\coloneqq \omega-t(\omega-\omega^\star)$. Like in \cref{proof_app:pfe_stability}, it follows:
		  \begin{align}
		  	\delta(\omega,\omega)&=\delta(\omega,\omega)-\underbrace{\delta(\omega^\star,\omega^\star)}_{=0},\\
		  	&=\delta\circ l(t=0) - \delta\circ l(t=1),\\
		  	&=-\int_0^1 \partial_t \delta\circ l(t) dt,\\ 
		  	&=\int_0^1\nabla_\omega \delta\circ l(t) dt(\omega-\omega^\star),
		  \end{align}
	  hence:
	  \begin{align}
	  	\nabla_\omega V(\omega)^\top \delta(\omega,\omega)&=(\omega-\omega^\star)^\top \int_0^1\nabla_\omega \delta\circ l(t) dt (\omega-\omega^\star),\\
	  	&=(\omega-\omega^\star)^\top\bar{J}_{TD}(\omega^\star,\omega)(\omega-\omega^\star),\\
	  	&<0,
	  \end{align}
 for all $\omega\in\mathcal{X}_{\textrm{TD}}(\omega^\star)$ under \cref{ass:TD_jacobian_stability}, 
as required. 
  	\end{proof}
\end{theorem}
\subsection{Stabilising FPE}

\begin{proposition}
Using the regularised TD vector in \cref{eq:regularised_update}, the path-mean Jacobians are:
	\begin{align}
		\bar{H}_\textrm{Reg}(\omega,\omega^\star;\bar{\omega}_l) =	\mu\bar{H}(\omega,\omega^\star;\bar{\omega}_l)-(1-\mu) \left(\bar{J}_{\delta}(\omega,\omega^\star;\bar{\omega}_l)- I\eta\right),\\
		\bar{J}_{\delta,\textrm{Reg}}(\omega,\omega^\star;\bar{\omega}_l)=\mu \bar{J}_{\delta}(\omega,\omega^\star;\bar{\omega}_l)-(1-\mu)\left(\bar{H}(\omega,\omega^\star;\bar{\omega}_l)- I\eta\right).
	\end{align}
	\cref{ass:FPE_stability} is satisfied if:
	\begin{align}
		\sup_{\omega,\omega'\in \mathcal{X}_{\textrm{FPE}}(\omega^\star)}\left\lVert  	\bar{H}_\textrm{Reg}(\omega',\omega^\star;\omega) ^{-1}	\bar{J}_{\delta,\textrm{Reg}}(\omega,\omega^\star;\omega^\star)\right\rVert < 1.\label{eq_app:regularised_condition}
	\end{align}
	There exists finite $\eta,\mu$ such that \cref{eq_app:regularised_condition} holds.
\begin{proof}
	Taking derivatives of $\delta_\textrm{Reg}(\omega,\omega')$:
	\begin{align}
		-\nabla_{\omega}\delta_\textrm{Reg}(\omega,\omega')&=-\mu\nabla_\omega\delta(\omega,\omega')-(1-\mu)\left(\nabla_\omega\delta(\omega',\omega)-I\eta\right),\\
		\implies \bar{H}_\textrm{Reg}(\omega,\omega^\star;\bar{\omega}_l)&=- \int^1_0 \nabla_{\omega'} \delta_\textrm{Reg}(\omega'=\omega-t(\omega-\omega^\star),\bar{\omega}_l)dt=\mu\bar{H}(\omega,\omega^\star;\bar{\omega}_l)-(1-\mu) \left(\bar{J}_{\delta}(\omega,\omega^\star;\bar{\omega}_l)- I\eta\right),\\
			\nabla_{\omega'}\delta_\textrm{Reg}(\omega,\omega')&=\mu\nabla_{\omega'}\delta(\omega,\omega')+(1-\mu)\left(\nabla_{\omega'}\delta(\omega',\omega)+I\eta\right),\\
		\implies \bar{J}_{\delta,\textrm{Reg}}(\omega,\omega^\star;\bar{\omega}_l)&= \int^1_0 \nabla_{\omega'} \delta_\textrm{Reg}(\bar{\omega}_l,\omega'=\omega-t(\omega-\omega^\star))dt= \mu \bar{J}_{\delta}(\omega,\omega^\star;\bar{\omega}_l)-(1-\mu)\left(\bar{H}(\omega,\omega^\star;\bar{\omega}_l)- I\eta\right).
	\end{align} 
For clarity, we drop arguments of $\omega',\omega^\star$ and $\omega$ from our notation. 
\begin{align}
 \bar{H}_\textrm{Reg} ^{-1}	\bar{J}_{\delta,\textrm{Reg}} &=\bar{H}_\textrm{Reg} ^{-1}	\left(\bar{J}_{\delta,\textrm{Reg}}-\bar{H}_\textrm{Reg} \right) + I,\\
 &=\left(\mu\bar{H}-(1-\mu)\left( \bar{J}_{\delta}-I\eta\right)\right)^{-1}	\left(\bar{J}_{\delta}-\bar{H} \right) + I,\\
  &=\left((2\mu -1 +(1-\mu)  )\bar{H}-(1-\mu)\left( \bar{J}_{\delta}- I\eta\right)\right)^{-1}	\left(\bar{J}_{\delta}-\bar{H} \right) + I,\\
    &=\left((2\mu -1 ) \bar{H}-(1-\mu) \left(\bar{J}_{\delta}- \bar{H}-I\eta\right)\right)^{-1}	\left(\bar{J}_{\delta}-\bar{H} \right) + I,\\
     &=\left((2\mu -1 ) \bar{H}-(1-\mu) \left(\bar{J}_{\delta}- \bar{H}-I\eta\right)\right)^{-1}	\left(\bar{J}_{\delta}-\bar{H} -\eta I +\eta I \right) + I,\\
       &=\left((2\mu -1 ) \bar{H}-(1-\mu) \left(\bar{J}_{\delta}- \bar{H}-I\eta\right)\right)^{-1}	\left(\bar{J}_{\delta}-\bar{H} -\eta I \right)\\
       &\qquad+ \left((2\mu -1 ) \bar{H}-(1-\mu) \left(\bar{J}_{\delta}- \bar{H}-I\eta\right)\right)^{-1}\eta I + I.
\end{align}  
We note that $\left(\bar{J}_{\delta}- \bar{H}-I\eta\right)$ can always be made non-singular (and hence invertible) through an arbitrarily small change in $\eta$, allowing us to multiply the first term  by $\left(\bar{J}_{\delta}- \bar{H}-I\eta\right)^{-1} \left(\bar{J}_{\delta}- \bar{H}-I\eta\right)=I$, yielding:
\begin{align}
	 \bar{H}_\textrm{Reg} ^{-1}	\bar{J}_{\delta,\textrm{Reg}} &=\left((2\mu -1 ) \bar{H} 	\left(\bar{J}_{\delta}-\bar{H} -\eta I \right)^{-1}-(1-\mu)I\right)^{-1}\\
	 &\qquad+ \left((2\mu -1 ) \bar{H}-(1-\mu) \left(\bar{J}_{\delta}- \bar{H}-I\eta\right)\right)^{-1}\eta I + I,\\
	 &=\left((2\mu -1 ) \bar{H} 	\left(\bar{J}_{\delta}-\bar{H} -\eta I \right)^{-1}-(1-\mu)I\right)^{-1}+ \left(\mu \bar{H}-(1-\mu) \bar{J}_{\delta}+(1-\mu)I\eta\right)^{-1}\eta I + I,
\end{align}
We observe that:
\begin{align}
	\lim_{\eta\rightarrow\infty}(2\mu -1 ) \bar{H} 	\left(\bar{J}_{\delta}-\bar{H} -\eta I \right)^{-1}=0,  \quad	\lim_{\eta\rightarrow\infty} \left(\mu \bar{H}-(1-\mu) \bar{J}_{\delta}+(1-\mu)I\eta\right)^{-1}\eta I=I \textrm{sgn}(1-\mu),
\end{align}
hence taking limits $\eta\rightarrow \infty$ yields:
\begin{align}
	\lim_{\eta\rightarrow\infty}\bar{H}_\textrm{Reg} ^{-1}	\bar{J}_{\delta,\textrm{Reg}} &= I\left(\textrm{sgn}(1-\mu)-\frac{1}{1-\mu}+1\right),\\
	&= I\left(\textrm{sgn}(1-\mu)-\frac{\mu}{1-\mu}\right).
\end{align}
From the continuity of the norm, it follows:
\begin{align}
\lim_{\eta\rightarrow\infty} \lVert 	\bar{H}_\textrm{Reg}(\omega,\omega^\star;\bar{\omega}_l) \rVert =\left\lvert \textrm{sgn}(1-\mu)-\frac{\mu}{1-\mu}\right\rvert,
\end{align}
implying that $\lim_{\eta\rightarrow\infty} \lVert 	\bar{H}_\textrm{Reg}(\omega,\omega^\star;\bar{\omega}_l) \rVert<1$ for any $\mu>2$, $\mu\in (0,\frac{2}{3})$, which it suffices assume for hereon out. From the definition of the limit, there exists some finite $\eta'$ such that 
\begin{align}
	\left\lVert  \bar{H}_\textrm{Reg} ^{-1}	\bar{J}_{\delta,\textrm{Reg}}\right\rVert \le1-\epsilon,
\end{align}
for all $\eta>\eta'$ for some small $0<\epsilon<1$, and hence
\begin{align}
		\left\lVert  \bar{H}_\textrm{Reg} ^{-1}	\bar{J}_{\delta,\textrm{Reg}}\right\rVert <1,
\end{align}
for all $\eta>\eta'$, as required.
\end{proof}
\end{proposition}

\subsection{Nonasymptotic Analysis}
\begin{lemma} \label{proof:factored_update}Under \cref{ass:regularity}, for $i>0$ the expected updates can be factored as:
	\begin{align}
		\mathbb{E}_{P_\varsigma}[\omega_{i+1}-\bar{\omega}_l^\star]&= \left(I-\alpha_l	\bar{H}(\omega_i,\bar{\omega}^\star_l;\bar{\omega}_l)\right)(\omega_i-\bar{\omega}^\star_l),\\
		\mathbb{E}_{P_\varsigma}[\omega_{i+1}-\omega^\star]&= \left(I-\alpha_l	\bar{H}(\omega_i,\bar{\omega}^\star_l;\bar{\omega}_l)\right)(\omega_i-\bar{\omega}^\star_l)+\bar{\omega}_l^\star-\omega^\star.
	\end{align}
and for $i=0$:
\begin{align}
		\mathbb{E}_{P_\varsigma}[\omega_{1}-\bar{\omega}_l^\star]=(I+\alpha \bar{J}_\textrm{TD}(\bar{\omega}_l,\omega^\star))(\bar{\omega}_l-\omega^\star)+\omega^\star-\bar{\omega}_l^\star
\end{align}
	\begin{proof}
		By the definition of the expected update $\omega_{i+1}$:
		\begin{align}
			\mathbb{E}_{P_\varsigma}[\omega_{i+1}-\bar{\omega}_l^\star]&=\omega_i-\bar{\omega}_l^\star+ \alpha_l\delta(\omega_i,\bar{\omega}_l)- \alpha_l\underbrace{\delta(\bar{\omega}_l^\star,\bar{\omega}_l)}_{=0}.
		\end{align}
		Like in \cref{proof_app:pfe_stability}, let $\ell(t)\coloneqq \omega_i-t(\omega_i-\bar{\omega}_l^\star)$ define the line connecting $\omega_i$ to $\bar{\omega}_l^\star$. Using this notation we re-write the expected update as:
		\begin{align}
			\mathbb{E}_{P_\varsigma}\left[\omega_{i+1}-\bar{\omega}^\star_l \right]=&\omega_i-\bar{\omega}_l^\star+ \alpha_l\left(\delta(\omega=\ell(0),\bar{\omega}_l)-\delta(\omega=\ell(1),\bar{\omega}_l) \right).
		\end{align}
		Applying the fundamental theorem of calculus under \cref{ass:regularity} and the chain rule yields our desired result:
		\begin{align}
			\mathbb{E}_{P_\varsigma}\left[\omega_{i+1}-\bar{\omega}^\star_l \right]&=\omega_i-\bar{\omega}_l^\star- \alpha_l\int^1_0\partial_t\delta(\omega=\ell(t),\bar{\omega}_l)dt,\\
			&=\omega_i-\bar{\omega}_l^\star- \alpha_l\int^1_0\nabla_\omega\delta(\omega,\bar{\omega}_l)_{\omega=\ell(t)}\partial_t\ell(t) dt,\\
			&=\omega_i-\bar{\omega}_l^\star+ \alpha_l\left(\int^1_0\nabla_\omega\delta(\omega,\bar{\omega}_l)_{\omega=\ell(t)} dt\right)(\omega_i-\bar{\omega}_l^\star),\\
			&= \left(I-\alpha_l	\bar{H}(\omega_i,\bar{\omega}^\star_l;\bar{\omega}_l)\right)(\omega_i-\bar{\omega}^\star_l).
		\end{align}
		Our second result follows immediately:
		\begin{align}
			\mathbb{E}_{P_\varsigma}[\omega_{i+1}-\omega^\star]&=\mathbb{E}_{P_\varsigma}[\omega_{i+1}-\bar{\omega}_l^\star] +\bar{\omega}_l^\star -\omega^\star ,\\
			&=\left(I-\alpha_l	\bar{H}(\omega_i,\bar{\omega}^\star_l;\bar{\omega}_l)\right)(\omega_i-\bar{\omega}^\star_l)+\bar{\omega}_l^\star-\omega^\star.
		\end{align}
	For our final result: 
	\begin{align}
				\mathbb{E}_{P_\varsigma}[\omega_{1}-\bar{\omega}_l^\star]&=\mathbb{E}_{P_\varsigma}[\omega_{1}-\omega^\star+\omega^\star-\bar{\omega}_l^\star],\\
				&=\mathbb{E}_{P_\varsigma}[\omega_{1}-\omega^\star]+\omega^\star-\bar{\omega}_l^\star.
	\end{align}
By the definition of the expected update:
		\begin{align}
	\mathbb{E}_{P_\varsigma}[\omega_{1}-\omega^\star]&=\bar{\omega}_l-\omega^\star+ \alpha_l\delta(\bar{\omega}_l,\bar{\omega}_l)- \alpha_l\underbrace{\delta(\omega^\star,\omega^\star)}_{=0}.
\end{align}
Let $\ell(t)\coloneqq \bar{\omega}_l-t(\bar{\omega}_l-\omega^\star)$ define the line connecting $\bar{\omega}_l$ to $\omega^\star$. Using this notation we re-write the expected update as:
\begin{align}
	\mathbb{E}_{P_\varsigma}[\omega_{1}-\omega^\star]=&\bar{\omega}_l-\omega^\star+ \alpha_l\left(\delta(\omega=\ell(0),\omega=\ell(t))-\delta(\omega=\ell(1),\omega=\ell(1)) \right).
\end{align}
Applying the fundamental theorem of calculus under \cref{ass:regularity} and the chain rule yields our desired result:
\begin{align}
	\mathbb{E}_{P_\varsigma}\left[\omega_{i+1}-\bar{\omega}^\star_l \right]&=\bar{\omega}_l-\omega^\star- \alpha_l\int^1_0\partial_t\delta(\omega=\ell(t),\omega=\ell(t))dt,\\
	&=\bar{\omega}_l-\omega^\star- \alpha_l\int^1_0\nabla_\omega\delta(\omega,\omega)\vert_{\omega=\ell(t)}\partial_t\ell(t) dt,\\
		&=\bar{\omega}_l-\omega^\star+ \alpha_l\left(\int^1_0\nabla_\omega\delta(\omega,\omega)\vert_{\omega=\ell(t)}dt\right)(\bar{\omega}_l-\omega^\star),\\
	&= \left(I+\alpha_l	\bar{J}_\textrm{TD}(\bar{\omega}^\star_l,\omega^\star)\right)(\omega_i-\bar{\omega}^\star_l).
\end{align}

	\end{proof}
\end{lemma}

\begin{lemma} \label{proof:norm_bound} Under \cref{ass:regularity},
	\begin{align}
		\mathbb{E}_{P_{\varsigma_i}}\left[\lVert\omega_{i+1}-\omega^\star\rVert\right] &\le \left\lvert 1-\alpha_l \lambda_H^\star	\right\rvert\left\lVert\omega_i-\bar{\omega}_l^\star\right\rVert+\left\lVert\bar{\omega}_l^\star-\omega^\star\right\rVert+\alpha_l\sigma_\delta.
	\end{align}		
	\begin{proof}
		We start by bounding the expected norm term using Jensen's inequality: $\mathbb{E}_X[\sqrt{X^2}]\le \sqrt{\mathbb{E}_X[X^2]}$:
		\begin{align}
			\mathbb{E}_{P_{\varsigma_{i}}}\left[	\left\lVert \omega_{i+1} -\omega^\star\right\rVert\right]&\le	  \sqrt{\mathbb{E}_{P_{\varsigma_{i}}}\left[	\left\lVert \omega_{i+1} -\omega^\star\right\rVert^2\right]},\\
			&=\sqrt{	\left\lVert \mathbb{E}_{P_{\varsigma_{i}}}\left[ \omega_{i+1} -\omega^\star\right]\right\rVert^2+\mathbb{V}_{P_{\varsigma_{i}}}\left[ \omega_{i+1} -\omega^\star\right]},\\
			&=\sqrt{	\left\lVert \mathbb{E}_{P_{\varsigma_{i}}}\left[ \omega_{i+1} -\omega^\star\right]\right\rVert^2+\mathbb{V}_{P_{\varsigma_{i}}}\left[ \omega_{i+1} \right]},\\
			&\le\left\lVert \mathbb{E}_{P_{\varsigma_{i}}}\left[ \omega_{i+1} -\omega^\star\right]\right\rVert+\sqrt{\mathbb{V}_{P_{\varsigma_{i}}}\left[ \omega_{i+1} \right]}
		\end{align}
		where we applied the triangle inequality to derive the final line. 	We bound the variance term by substituting $\omega_{i+1} =\omega_{i} +\alpha_l\delta(\omega_{i},\bar{\omega}_l,\varsigma_{i})$:
		\begin{align}
			\mathbb{V}_{P_{\varsigma_{i}}}\left[\omega_{i+1}\right]&=(\alpha_l)^2\mathbb{E}_{P_{\varsigma_{i}}}\left[\left\lVert\delta(\omega_{i},\bar{\omega}_l,\varsigma_{i})-\mathbb{E}_{P_{\varsigma_{i}}}\left[\delta(\omega_{i},\bar{\omega}_l,\varsigma_{i})\right]\right\rVert^2\right],\\
			&=(\alpha_l)^2\mathbb{V}_{P_{\varsigma_{i}}}\left[\delta(\omega_{i},\bar{\omega}_l,\varsigma_{i})\right],\\
			&\le(\alpha_l\sigma_\delta)^2,\\
			\implies  \mathbb{E}_{P_{\varsigma_{i}}}\left[	\left\lVert \omega_{i+1} -\omega^\star\right\rVert\right]&\le \left\lVert \mathbb{E}_{P_{\varsigma_{i}}}\left[ \omega_{i+1} -\omega^\star\right]\right\rVert+\alpha_l\sigma_\delta\label{eq:variance_bound}
		\end{align}
		Applying \cref{proof:factored_update} to the expectation and using the triangle inequality yields our desired result:
		\begin{align}
			\mathbb{E}_{P_{\varsigma_{i}}}\left[	\left\lVert \omega_{i+1} -\omega^\star\right\rVert\right]&\le \left\lVert\left(I-\alpha_l	\bar{H}(\omega_i,\bar{\omega}^\star_l;\bar{\omega}_l)\right)(\omega_i-\bar{\omega}_l^\star)+\left(\bar{\omega}_l^\star-\omega^\star\right)\right\rVert+\alpha_l\sigma_\delta,\\
			&\le \left\lVert  I-\alpha_l	\bar{H}(\omega_i,\bar{\omega}^\star_l;\bar{\omega}_l)\right\rVert \left\lVert\omega_i-\bar{\omega}_l^\star\right\rVert+\left\lVert\bar{\omega}_l^\star-\omega^\star\right\rVert+\alpha_l\sigma_\delta,\\
			&\le \sup_{\omega_i,\bar{\omega}^\star_l,\bar{\omega}_l}\left\lVert  I-\alpha_l	\bar{H}(\omega_i,\bar{\omega}^\star_l;\bar{\omega}_l)\right\rVert \left\lVert\omega_i-\bar{\omega}_l^\star\right\rVert+\left\lVert\bar{\omega}_l^\star-\omega^\star\right\rVert+\alpha_l\sigma_\delta,\\
			&= \left\lvert 1-\alpha_l \lambda_H^\star	\right\rvert\left\lVert\omega_i-\bar{\omega}_l^\star\right\rVert+\left\lVert\bar{\omega}_l^\star-\omega^\star\right\rVert+\alpha_l\sigma_\delta.
		\end{align}
		
	\end{proof}
\end{lemma}

\begin{theorem}\label{proof_app:one_step_bound}
	Define 
	\begin{align}
		\sigma_k&\coloneqq
		\left(1-\left\lvert 1-\alpha_l \lambda_H^\star	\right\rvert^k\right)\frac{\sigma_\delta}{\lambda_H^\star},
	\end{align}
	Let Assumptions~\ref{ass:iid} and~\ref{ass:regularity} hold, then:
	\begin{align}
		\mathbb{E}\left[\lVert\bar{\omega}_{l+1}-\omega^\star\rVert \right]\le \mathcal{C}(\alpha_l,k)\mathbb{E}\left[\lVert\bar{\omega}_l-\omega^\star\rVert\right]+\alpha_l\sigma_k.
	\end{align}
	\begin{proof}
		Let $\{\omega_i\}_{i=0}^k$ denote the intermediate function approximation parameters between target parameter updates $\bar{\omega}_{l+1}$ and $\bar{\omega}_l$, with $\omega_0=\bar{\omega}_l$ and $\omega_k=\bar{\omega}_{l+1}$. We define the set of samples up to $i$ as: $\mathcal{D}_i\coloneqq\{\varsigma_j\}_{j=0}^i$ with distribution $P_{\mathcal{D}_i}$, with sample $\varsigma_j$ having distribution $P_{\varsigma_j}$. Under this notation, we must show:
		\begin{align}
			\mathbb{E}_{P_{\mathcal{D}_{k-1}}}\left[\lVert\omega_k-\omega^\star\rVert \right]\le \mathcal{C}(\alpha_l,k)\lVert\omega_0-\omega^\star\rVert+\alpha_l\sigma_k.\label{eq:expected_difference_notation}
		\end{align}
		Applying \cref{proof:norm_bound} to the inner expectation:
		\begin{align}
			\mathbb{E}_{P_{\mathcal{D}_{k-1}}}\left[\lVert\omega_k-\omega^\star\rVert \right]&=		\mathbb{E}_{P_{\mathcal{D}_{k-2}}}\left[\mathbb{E}_{P_{\varsigma_{k-1}}}\left[\lVert\omega_k-\omega^\star\rVert \right]\right],\\
			&\le	\mathbb{E}_{P_{\mathcal{D}_{k-2}}}\left[\left\lvert 1-\alpha_l \lambda_H^\star	\right\rvert\left\lVert\omega_{k-1}-\bar{\omega}_l^\star\right\rVert+\left\lVert\bar{\omega}_l^\star-\omega^\star\right\rVert+\alpha_l\sigma_\delta\right],\\
			&=	\left\lvert 1-\alpha_l \lambda_H^\star	\right\rvert\mathbb{E}_{P_{\mathcal{D}_{k-2}}}\left[\left\lVert\omega_{k-1}-\bar{\omega}_l^\star\right\rVert\right]+\left\lVert\bar{\omega}_l^\star-\omega^\star\right\rVert+\alpha_l\sigma_\delta,\\
			&=	\left\lvert 1-\alpha_l \lambda_H^\star	\right\rvert\mathbb{E}_{P_{\mathcal{D}_{k-3}}}\left[\mathbb{E}_{P_{\varsigma_{k-2}}}\left[\left\lVert\omega_{k-1}-\bar{\omega}_l^\star\right\rVert\right]\right]+\left\lVert\bar{\omega}_l^\star-\omega^\star\right\rVert+\alpha_l\sigma_\delta.\label{eq:first_bound}
		\end{align}
		Applying \cref{eq:variance_bound} from \cref{proof:norm_bound} to the inner expectation and applying \cref{proof:factored_update} yields:
		\begin{align}
			\mathbb{E}_{P_{\varsigma_{k-2}}}\left[\left\lVert\omega_{k-1}-\bar{\omega}_l^\star\right\rVert\right]&\le \left\lVert \mathbb{E}_{P_{\varsigma_{k-2}}}\left[ \omega_{k-1} -\omega^\star\right]\right\rVert+\alpha_l\sigma_\delta,\\
			&\le \left\lVert	\left(I-\alpha_l	\bar{H}(\omega_{k-2},\bar{\omega}^\star_l;\bar{\omega}_l)\right)(\omega_{k-2}-\bar{\omega}^\star_l)\right\rVert+\alpha_l\sigma_\delta,\\
			&\le  \sup_{\omega_{k-2},\bar{\omega}^\star_l,\bar{\omega}_l}\left\lVert  I-\alpha_l	\bar{H}(\omega_{k-2},\bar{\omega}^\star_l;\bar{\omega}_l)\right\rVert \left\lVert\omega_{k-2}-\bar{\omega}_l^\star\right\rVert+\alpha_l\sigma_\delta,\\
			&= \left\lvert  I-\alpha_l	\lambda_H^\star\right\rvert \left\lVert\omega_{k-2}-\bar{\omega}_l^\star\right\rVert+\alpha_l\sigma_\delta. \label{eq:recursive_bound}
		\end{align}
		Recursively applying \cref{eq:recursive_bound} to \cref{eq:first_bound} $k-1$ times yields:
		\begin{align}
			\mathbb{E}_{P_{\mathcal{D}_{k-1}}}\left[\lVert\omega_k-\omega^\star\rVert \right]&\le			\mathbb{E}_{P_{\varsigma_0}}\left[\left\lvert 1-\alpha_l \lambda_H^\star	\right\rvert^{k-1}\left\lVert\omega_1-\bar{\omega}_l^\star\right\rVert\right]+\left\lVert\bar{\omega}_l^\star-\omega^\star\right\rVert+\sum_{i=0}^{k-2} \left\lvert 1-\alpha_l \lambda_H^\star	\right\rvert^i\alpha_l\sigma_\delta,\\
&=	\left\lvert 1-\alpha_l \lambda_H^\star	\right\rvert^{k-1}	\mathbb{E}_{P_{\varsigma_0}}\left[\left\lVert\omega_1-\omega^\star_l\right\rVert\right]+\left\lVert\bar{\omega}_l^\star-\omega^\star\right\rVert+\sum_{i=0}^{k-2} \left\lvert 1-\alpha_l \lambda_H^\star	\right\rvert^i\alpha_l\sigma_\delta.\label{eq:recursive_bound}
\end{align}
Now, applying \cref{eq:variance_bound} and \cref{proof:factored_update} to the expectation:
\begin{align}
	\mathbb{E}_{P_{\varsigma_0}}\left[\left\lVert\omega_1-\omega^\star_l\right\rVert\right]&\le 	\left\lVert\mathbb{E}_{P_{\varsigma_0}}\left[\omega_1-\omega^\star_l\right]\right\rVert+\alpha_l\sigma_\delta,\\
	&= 	\left\lVert(I+\alpha \bar{J}_\textrm{TD}(\bar{\omega}_l,\omega^\star))(\bar{\omega}_l-\omega^\star)+\omega^\star-\bar{\omega}_l^\star\right\rVert+\alpha_l\sigma_\delta,\\
	&\le  \left\lVert I+\alpha \bar{J}_\textrm{TD}(\bar{\omega}_l,\omega^\star)\right\rVert\lVert\bar{\omega}_l-\omega^\star\rVert+\lVert\bar{\omega}_l^\star-\omega^\star\rVert+\alpha_l\sigma_\delta,\\
		&=\left\lVert\bar{J}_\textrm{TD}^\star\right\rVert\lVert\bar{\omega}_l-\omega^\star\rVert+\lVert\omega^\star_l-\bar{\omega}_l\rVert+\alpha_l\sigma_\delta.
\end{align}
Substituting into \cref{eq:recursive_bound}:
		\begin{align}
	\mathbb{E}_{P_{\mathcal{D}_{k-1}}}&\left[\lVert\omega_k-\omega^\star\rVert \right]\le		\left\lvert 1-\alpha_l \lambda_H^\star	\right\rvert^{k-1}\left\lVert\bar{J}_\textrm{TD}^\star\right\rVert\left\lVert\bar{\omega}_l-\omega^\star\right\rVert+(1+\left\lvert 1-\alpha_l \lambda_H^\star	\right\rvert^{k-1})\left\lVert\bar{\omega}_l^\star-\omega^\star\right\rVert+\sum_{i=0}^{k-1} \left\lvert 1-\alpha_l \lambda_H^\star	\right\rvert^i\alpha_l\sigma_\delta.\\
			&=	\left\lvert 1-\alpha_l \lambda_H^\star	\right\rvert^{k-1}\left\lVert\bar{J}_\textrm{TD}^\star\right\rVert\left\lVert\bar{\omega}_l-\omega^\star\right\rVert+(1+\left\lvert 1-\alpha_l \lambda_H^\star	\right\rvert^{k-1})\left\lVert\bar{\omega}_l^\star-\omega^\star\right\rVert+\frac{ 1-\left\lvert 1-\alpha_l \lambda_H^\star	\right\rvert^k}{1-\left\lvert 1-\alpha_l \lambda_H^\star	\right\rvert}\alpha_l\sigma_\delta,\\
			&=	\left\lvert 1-\alpha_l \lambda_H^\star	\right\rvert^{k-1}\left\lVert\bar{J}_\textrm{TD}^\star\right\rVert\left\lVert\bar{\omega}_l-\omega^\star\right\rVert+(1+\left\lvert 1-\alpha_l \lambda_H^\star	\right\rvert^{k-1})\left\lVert\bar{\omega}_l^\star-\omega^\star\right\rVert+\left(1-\left\lvert 1-\alpha_l \lambda_H^\star	\right\rvert^k\right)\frac{\sigma_\delta}{\lambda_H^\star},\\
			&\le\left\lvert 1-\alpha_l \lambda_H^\star	\right\rvert^{k-1}\left\lVert\bar{J}_\textrm{TD}^\star\right\rVert\left\lVert\bar{\omega}_l-\omega^\star\right\rVert+(1+\left\lvert 1-\alpha_l \lambda_H^\star	\right\rvert^{k-1})\left\lVert\bar{\omega}_l^\star-\omega^\star\right\rVert+\sigma_k.
		\end{align}
		Finally, we apply \cref{proof_app:pfe_stability} to yield our desired result:
		\begin{align}
			\mathbb{E}_{P_{\mathcal{D}_{k-1}}}&\left[\lVert\omega_k-\omega^\star\rVert \right]\\
			&\le	\left\lvert 1-\alpha_l \lambda_H^\star	\right\rvert^{k-1}\left\lVert\bar{J}_\textrm{TD}^\star\right\rVert\left\lVert\bar{\omega}_l-\omega^\star\right\rVert+\left(1+	\left\lvert 1-\alpha_l \lambda_H^\star	\right\rvert^{k-1}\right)\left\lVert\bar{H}(\bar{\omega}_l^\star,\omega^\star;\bar{\omega}_{l})^{-1} \bar{J}_\delta(\bar{\omega}_{l},\omega^\star;\omega^\star)(\bar{\omega}_{l}-\omega^\star)\right\rVert+\sigma_k,\\
			&\le\left\lvert 1-\alpha_l \lambda_H^\star	\right\rvert^{k-1}\left\lVert\bar{J}_\textrm{TD}^\star\right\rVert\left\lVert\bar{\omega}_l-\omega^\star\right\rVert+\left(1+	\left\lvert 1-\alpha_l \lambda_H^\star	\right\rvert^{k-1}\right)\left\lVert \bar{H}(\bar{\omega}_l^\star,\omega^\star;\bar{\omega}_{l})^{-1} \bar{J}_\delta(\bar{\omega}_{l},\omega^\star;\omega^\star)\right\rVert\left\lVert\bar{\omega}_l-\omega^\star\right\rVert+\sigma_k,\\
			&\le	\left\lvert 1-\alpha_l \lambda_H^\star	\right\rvert^{k-1}\left\lVert\bar{J}_\textrm{TD}^\star\right\rVert\left\lVert\bar{\omega}_l-\omega^\star\right\rVert+\left(1+	\left\lvert 1-\alpha_l \lambda_H^\star	\right\rvert^{k-1}\right)\left\lVert \bar{J}_\textrm{FPE}^\star\right\rVert\left\lVert\bar{\omega}_l-\omega^\star\right\rVert+\sigma_k,\\
			&=	\mathcal{C}(\alpha_l,k)\left\lVert\bar{\omega}_l-\omega^\star\right\rVert+\sigma_k.
		\end{align}
	\end{proof}
\end{theorem}

\begin{corollary}\label{proof_app:nonasymptotic} Let Assumptions~\ref{ass:iid},~\ref{ass:regularity},~\ref{ass:FPE_stability} and~\ref{ass:contraction_region} hold.  For a fixed stepsize  $\alpha_l=\alpha>0$.  For a fixed stepsize  $\alpha_l=\alpha>0$,
\begin{align}
	&\mathbb{E}\left[
	\lVert \bar{\omega}_l- \omega^\star\rVert
	\right]
	\le \frac{\alpha\sigma_k}{1-c}+\exp(-l(1-c))\left(
	\left\lVert
	\bar{\omega}_0- \omega^\star
	\right\rVert
	-\frac{\sigma_k}{1-c}
	\right)	.
\end{align}
\begin{proof}
We start by applying \cref{proof_app:one_step_bound}:	
	\begin{align}
			\mathbb{E}\left[\lVert\bar{\omega}_l-\omega^\star\rVert \right]\le \mathcal{C}(\alpha_l,k)\mathbb{E}\left[\lVert\bar{\omega}_{l-1}-\omega^\star\rVert\right]+\alpha_l\sigma_k.
	\end{align}
As $\mathcal{X}_\textrm{FPE}(\omega^\star)$ is a region of contraction and $\bar{\omega}_l\in\mathcal{X}_\textrm{FPE}(\omega^\star)$ for all $l\ge0$, there exists a positive $c<1$ under \cref{ass:contraction_region} such that $\mathcal{C}(\alpha_l,k)\le c$, hence:
	\begin{align}
		\mathbb{E}\left[\lVert\bar{\omega}_l-\omega^\star\rVert \right]\le c\mathbb{E}\left[\lVert\bar{\omega}_{l-1}-\omega^\star\rVert\right]+\alpha_l\sigma_k.\label{eq:single_step_bound}
\end{align}
Now, for a fixed constant stepsize $\alpha_l=\alpha$, we can apply \cref{eq:single_step_bound} $l$ times, yielding:
\begin{align}
	\mathbb{E}\left[	\lVert \bar{\omega}_l- \omega^\star \rVert\right]& \le c^{l}\left\lVert\bar{\omega}_0- \omega^\star \right\rVert+\alpha \sigma_k\sum_{i=0}^{l-1} c^{i},\\
& = c^{l}\left\lVert\bar{\omega}_0- \omega^\star \right\rVert+\alpha \sigma_k\frac{1-c^{l}}{1-c}\\
& = c^{l}\left(\left\lVert\bar{\omega}_0- \omega^\star \right\rVert-\frac{\alpha \sigma_k}{1-c}\right)+\frac{\alpha \sigma_k}{1-c},\\
& = (1-(1-c))^{l}\left(\left\lVert\bar{\omega}_0- \omega^\star \right\rVert-\frac{\alpha \sigma_k}{1-c}\right)+\frac{\alpha \sigma_k}{1-c}.
\end{align}
Now we apply the bound $1-x\le \exp(-x)$, yielding our desired result: 
\begin{align}
	\mathbb{E}\left[	\lVert \bar{\omega}_l- \omega^\star \rVert \right]& \le\exp(-(1-c))^{l}\left(\left\lVert\bar{\omega}_0- \omega^\star \right\rVert-\frac{(\alpha \sigma_k)}{1-c}\right)+\frac{\alpha \sigma_k}{1-c},\\
	&=\exp(-l(1-c))\left(\left\lVert\bar{\omega}_0- \omega^\star \right\rVert-\frac{\alpha \sigma_k}{1-c}\right)+\frac{\alpha \sigma_k}{1-c}.
\end{align} 
\end{proof}
\end{corollary}

\subsection{Breaking the Deadly Triad}
\begin{theorem} \label{proof_app:convergent_stepizes}Let \cref{ass:eigenvalues} hold over $\mathcal{X}_\textrm{FPE}(\omega^\star)$ from \cref{def:condition_function}. 
	For any $\frac{1}{\alpha_l}>\frac{\lambda_1^{\min}+\lambda_1^{\max}}{2}$ such that $\alpha_l>0$, any
	\begin{align}
		k> 1+\frac{\log(1-\lVert\bar{J}_\textrm{FPE}^\star\rVert)-\log(\lVert \bar{J}_\textrm{TD}^\star\rVert+\lVert\bar{J}_\textrm{FPE}^\star\rVert) }{\log(1-\alpha \lambda^\textrm{min})},
	\end{align}
	ensures that $\mathcal{X}_\textrm{FPE}(\omega^\star)$ is a region of contraction satisfying \cref{ass:contraction_region}.
	\begin{proof}
	Now, as $\left\lvert 1-\alpha_l \lambda' \right\rvert$ is a symmetric function of $\lambda$ with a minima at $\lambda=\frac{1}{\alpha_l}$ and $\frac{\lambda_1^{\min}+\lambda_1^{\max}}{2}$ is the mid point of $\lambda_1^{\min}$ and $\lambda_1^{\max}$, it follows: 
		  \begin{align}
		  		\lambda_H^\star\coloneqq \sup_{\omega,\omega'\in \mathcal{X}_\textrm{FPE}(\omega^\star)}\argsup_{ \lambda'\in\lambda(\nabla^2_\omega\mathcal{L}(\omega,\omega'))}\left\lvert 1-\alpha_l \lambda' \right\rvert=\lambda_1^{\min}.
		  \end{align}
	  Now,
	  \begin{gather}
	  	\alpha_l<\frac{2}{\lambda_1^{\min}+\lambda_1^{\max}}\implies\lambda^\star_H \le \frac{2}{\alpha_l}\implies\left\lvert 1-\alpha_l \lambda^\star_H\right\rvert<1,
	  \end{gather}
hence
\begin{gather}
	\lim_{k\rightarrow\infty}\mathcal{C}(\alpha_l,k)=	\lim_{k\rightarrow\infty} \left\lvert 1-\alpha_l \lambda^\textrm{min}	\right\rvert^{k-1}\lVert J_\textrm{TD}^\star\rVert+	\lim_{k\rightarrow\infty}\left(1+	\left\lvert 1-\alpha_l \lambda^\textrm{min}	\right\rvert^{k-1}\right)\left\lVert \bar{J}_\textrm{FPE}^\star\right\rVert=\left\lVert \bar{J}_\textrm{FPE}^\star\right\rVert<1.
\end{gather}
Let $\left\lVert \bar{J}_\textrm{FPE}^\star\right\rVert=1-\epsilon$  where $0<\epsilon<1$. From the definition of a limit, this implies that for $\epsilon$ there exists some finite $k'$ such that whenever $k>k'$:
\begin{align}
	\left\lvert \mathcal{C}(\alpha_l,k)-\left\lVert \bar{J}_\textrm{FPE}^\star\right\rVert\right\rvert <\epsilon\implies \lvert \mathcal{C}(\alpha_l,k)-(1-\epsilon)\rvert <\epsilon\implies\mathcal{C}(\alpha_l,k)<1,
\end{align}
as required. To find the value of $k$ for which $\mathcal{C}(\alpha_l,k)<1$, we set $\mathcal{C}(\alpha_l,k)=1$ and solve:
\begin{align}
	1&=	\left\lvert 1-\alpha_l \lambda^\textrm{min}	\right\rvert^{k-1}\left\lVert\bar{J}_\textrm{TD}^\star\right\rVert+\left(1+	\left\lvert 1-\alpha_l \lambda^\textrm{min}	\right\rvert^{k-1}\right)\left\lVert \bar{J}_\textrm{FPE}^\star\right\rVert,\\
\implies	\left\lvert 1-\alpha_l \lambda^\textrm{min}	\right\rvert^{k-1}&=	\frac{1-\left\lVert \bar{J}_\textrm{FPE}^\star\right\rVert}{\lVert \bar{J}_\textrm{TD}^\star\rVert+\left\lVert \bar{J}_\textrm{FPE}^\star\right\rVert },\\
		\implies  (k-1)\log(\left\lvert 1-\alpha_l \lambda^\textrm{min}	\right\rvert)&= \log(	1-\left\lVert \bar{J}_\textrm{FPE}^\star\right\rVert)- \log(	\lVert \bar{J}_\textrm{TD}^\star\rVert+\left\lVert \bar{J}_\textrm{FPE}^\star\right\rVert),\\
			\implies k&= 1+\frac{\log(1-\lVert\bar{J}_\textrm{FPE}^\star\rVert)-\log(\lVert \bar{J}_\textrm{TD}^\star\rVert+\lVert\bar{J}_\textrm{FPE}^\star\rVert) }{\log(1-\alpha \lambda^\textrm{min})}.
\end{align}
	\end{proof}
\end{theorem}

%% file: Appendix/appendix_experiments.tex
\section{Additional Experiment Information}
\label{app:experiments}
For both plots, each configuration was run over $5$ random seeds, with the central tendency given by the mean, and the shaded errors representing the standard error of the mean. Hyperparameters that are not varied in the plots were optimised by grid search across either linear or logarithmic hyperparameter ranges, as is suitable. Parameters were chosen that led to the highest performance as averaged across random seeds, then relevant hyperparameters were varied, using the optimal fixed hyperparameters. Hyperparameters that were varied are denoted as lists in the tables below.
\subsection{Baird's Counterexample}
\begin{wrapfigure}{R}{0.5\textwidth}
	\centering
	\includegraphics[width=\linewidth]{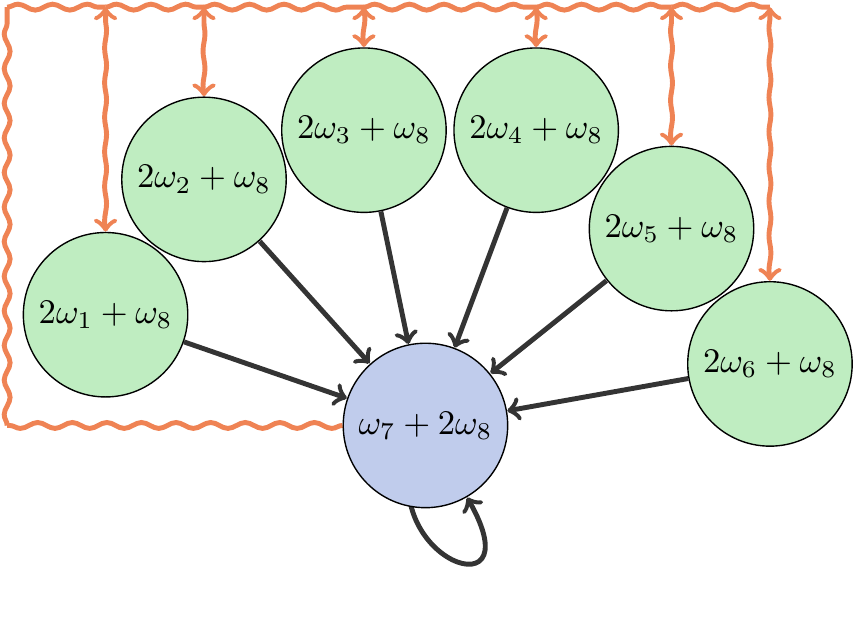}
	\caption{Baird's Counterexample. The solid (grey) action moves the agent to the lower state deterministically. The wavy (orange) action puts the agent into one of the upper states with equal probability}
	\vspace{-2cm}
	\label{fig:baird_env}
\end{wrapfigure}
Figure \cref{fig:baird_env} shows the counterexample. The behaviour policy chooses between the action represented by the wavy line with probability $6/7$, and the solid line with probability 1/7. The behaviour policy always chooses the solid line. The linear function approximation scheme is shown in terms of the value function weights. Sampling off policy in this way leads to divergence of TD, but PFPE converges, as seen in \cref{fig:baird_loss}.

\subsection{Cartpole Experiment}
For the Cartpole experiment, we use a simple DQN-style setup with a small multilayer perceptron (MLP) representing the value function. A small adjustment is made from PFPE as characterised by the paper. Instead of updating value parameters on single data points, parameter updates are averaged across a small batch. This was found to increase stability of learning in both settings, with no notable effects when comparing across independent variables. This means that, in addition to our target network, we also make use of a replay buffer which stores observed transitions. As such, data used in updates was sampled uniformly from previous transitions. The policy was $\epsilon$-greedy, with the estimated optimal action taken with probability $1 - \epsilon$. The environment is maintained by OpenAI as part of the gym suite, and falls under MIT licensing.
\begin{table}[H]
	\vspace{2cm}
\centering
\begin{tabular}{| l | c |}
\hline
Parameter & Value\\
\hline
\textbf{Environment Parameters}&\\
$\gamma$ & 0.99 \\
\textbf{Architecture Parameters}&\\
MLP Hidden Layers & 2 \\
Hidden Layer Size & 32 \\
Nonlinearity & ReLU \\
$\epsilon$ & 0.05 \\
\textbf{Training Parameters}&\\
Total Target Network Updates & 500 \\
Learning Rate & [0.001, 0.0005] \\
Momentum ($\mu$) & [0, 0.01] \\
Batch Size & 500 \\
Steps per Target Network Update ($k$) & 5\\
Data Gathering Steps per Update & 5 \\
Replay Buffer Size & 2500 \\
\hline
\end{tabular}
\caption{Relevant Parameters for Cartpole Experiment}
\label{tab:Cartpole}
\end{table}

\section{Extensions}
\label{app:extensions}
As discussed in \cref{sec:asymptotic_pfpe}, once we can establish \cref{ass:TD_jacobian_stability} then there are several theoretical tools that become applicable from stochastic approximation to prove convergence under a range of assumptions.
\citet{Kushner06} provide a comprehensive overview of classic methods. In particular, stochastic approximation has been shown to converge when sampling from an ergodic Markov chain under specific regularity assumptions \citep{kuhn10}. Perhaps the easiest to verify in our context is those of 
 \citet{andrieu05}, who provides a series of assumptions that can be checked in practice. Moreover, this theory was recently extended to Markov chains that converge sub-geometrically to their station distributions by \citet{debavelaere21}. Adherence of the updates to remain in a contractive region can be ensured by projection into an ever increasing subset of $\Omega$ until convergence occurs, which is detailed and analysed in \citet{Andradottir91}.